\pdfoutput=1

\documentclass[11pt]{article}

\usepackage[]{EACL2023}

\usepackage{times}
\usepackage{latexsym}

\usepackage[T1]{fontenc}

\usepackage[utf8]{inputenc}

\usepackage{microtype}

\usepackage{inconsolata}

\usepackage{graphicx}
\usepackage{eurosym}
\usepackage{hhline}
\usepackage{tablefootnote}
\usepackage{kantlipsum}
\usepackage{tabularx}
\graphicspath{ {figures/} }
\usepackage{hyperref}
\usepackage{amsmath}
\usepackage{hhline}
\usepackage[utf8]{inputenc}
\usepackage{multirow}

\usepackage[T1]{fontenc}
\usepackage{kantlipsum}
\usepackage{listings}
\usepackage{stfloats}
\usepackage{csquotes}
\usepackage{array}
\usepackage{booktabs}
\usepackage[super]{nth}
\usepackage[inline,shortlabels]{enumitem}
\usepackage{enumitem}
\usepackage{makecell}
\usepackage{siunitx}
\usepackage{xcolor, colortbl}
\usepackage{bm}
\usepackage{url}
\usepackage{wrapfig}
\usepackage{amsmath}
\usepackage{amssymb}
\usepackage{mathtools}
\usepackage{amsfonts}
\usepackage{enumitem}
\usepackage{graphicx}
\usepackage{subcaption}
\usepackage{nicefrac}       
\usepackage{tablefootnote}
\usepackage{bbm}
\usepackage{microtype} 
\usepackage{dsfont}
\usepackage{adjustbox}

\usepackage{algorithm,lipsum,changepage}

\newcommand{\dtemplama}{\textsc{DynamicTempLAMA}}
\newcommand{\templama}{\textsc{TempLAMA}}

\newcommand{\timelms}{\textsc{TimeLMs}}
\newcommand{\roberta}{\textsc{RoBERTa}}

\usepackage{relsize}

\newcommand{\Dnew}{\mathcal{D}_{t+1}^{\textsc{new}}}
\newcommand{\Dun}{\mathcal{D}_{t+1}^{\textsc{unchanged}}}
\newcommand{\Dup}{\mathcal{D}_{t+1}^{\textsc{updated}}}
\newcommand{\Ddel}{\mathcal{D}_{t+1}^{\textsc{deleted}}}
\title{Dynamic Benchmarking of Masked Language Models \\ on Temporal Concept Drift with Multiple Views}

\author{
Katerina Margatina$^{\diamondsuit}$\thanks{\, \, Work done during an internship at AWS AI Labs.}\quad Shuai Wang${^\dagger}$ \quad Yogarshi Vyas$^\dagger$ \\ 
{\bf Neha Anna John${^\dagger}$ \quad Yassine Benajiba${^\dagger}$ \quad Miguel Ballesteros${^\dagger}$ }\\
$^\diamondsuit$University of Sheffield\quad$^\dagger$AWS AI Labs\\
\texttt{k.margatina@sheffield.ac.uk},\\ \texttt{\{wshui,yogarshi,nehajohn,benajiy,ballemig\}@amazon.com}\\
  }

\begin{document}
\maketitle
\begin{abstract}
\textit{Temporal concept drift} refers to the problem of data changing over time. In NLP, that would entail that \textit{language} (e.g. new expressions, meaning shifts) and \textit{factual knowledge} (e.g. new concepts, updated facts) evolve over time. Focusing on the latter, we benchmark $11$ pretrained masked language models (MLMs) on a series of tests designed to evaluate the effect of temporal concept drift, as it is crucial that widely used language models remain up-to-date with the ever-evolving factual updates of the real world. Specifically, we provide a holistic framework that (1) \textit{dynamically} creates temporal test sets of any time granularity (e.g. month, quarter, year) of factual data from Wikidata, (2) constructs fine-grained splits of tests (e.g. updated, new, unchanged facts) to ensure \textit{comprehensive} analysis, and (3) evaluates MLMs in three distinct ways (single-token probing, multi-token generation, MLM scoring). In contrast to prior work, our framework aims to unveil how \textit{robust} an MLM is over time and thus to provide a signal in case it has become \textit{outdated}, by leveraging \textit{multiple views of evaluation}.
\end{abstract}

\section{Introduction}\label{sec:intro}
In the real world, what people talk about and how they tend to speak and write changes constantly over time. In Natural Language Processing (NLP), this entails a challenging shift of the textual data distribution that is commonly referred to as \textit{temporal concept drift}. Prior work has identified that pretrained language models (PLMs) tend to become outdated soon after new topics and concepts are emerging~\cite{lazaridou2021mind,dhingra-etal-2022-time,Agarwal2022TemporalEO,luu-etal-2022-time}, limiting their capability to be robust to newly generated data. 

We consider the desiderata of language models' robustness to temporal drift to be twofold. First, LMs should be well adapted to the dynamic use of language, from the \textit{linguistic} perspective. Language changes over time, pronunciations evolve, new words and expressions are borrowed or invented, the meaning of old words drifts, and morphology develops or decays~\cite{Blank1999WhyDN,traugott_dasher_2001,10.1145/2736277.2741627}.
Second, LMs should be aware of the ever-changing reality of the world, from a \textit{factual} perspective. Models' factual knowledge should be up-to-date with new facts and concepts (e.g. Covid-$19$) to be of use continuously. In this work, we focus on the latter; \textit{the temporal robustness of LMs to facts that change over time}.

\begin{figure}[!t]
        \centering
        \includegraphics[width=\columnwidth]{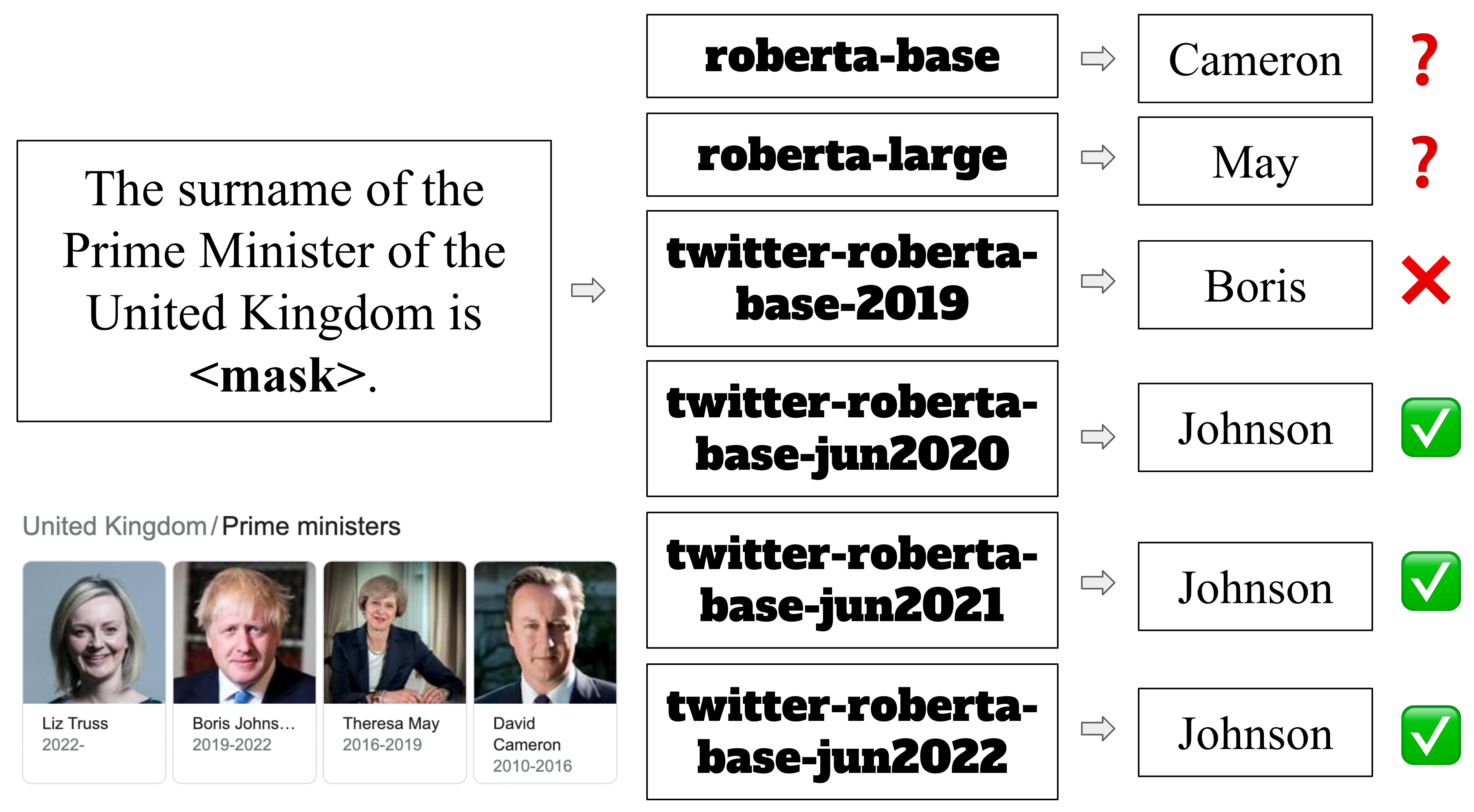}

    \caption{Querying pretrained MLMs on their knowledge about the Prime Minister of the United Kingdom.
    }
    \label{fig:intro}
\end{figure}

In an ideal scenario, we would like to know exactly when the factual knowledge of a model is ``expired'' so that we could adapt it to the new (or updated) set of facts. In reality, this is a challenging task. A large body of work has focused on the part of (continually) \textit{adapting} an ``outdated'' model to the new data distribution ~\cite{10.5555/3524938.3525306, yogatama-etal-2021-adaptive,sun2020lamal,biesialska-etal-2020-continual,jang2022towards, jin-etal-2022-lifelong, chakrabarty2022finetuned}. This line of work is parallel to ours, as we focus on the crucial step before adaptation, the evaluation of the model on temporal concept drift: \textit{How can we know if a language model is outdated or not?}

Let us consider the case  where we desire a language model to be up-to-date with the Prime Minister of the United Kingdom (Figure \ref{fig:intro}).\footnote{The time of writing of this paper is September $2022$.} A plausible way to evaluate this is to use the \textsc{lama}-probe paradigm~\cite{petroni-etal-2019-language} and query the LM as a knowledge base (KB). This would mean that we could form the query as \texttt{``The surname of the Prime Minister of the United Kingdom is <mask>.''}, give it as an input to a (masked) LM and inspect the output token distribution for the \texttt{<mask>} token.
Figure~\ref{fig:intro} shows the top prediction for a series of \textsc{RoBERTa} models.\footnote{Except for the \textsc{RoBERTa} \texttt{base} and \texttt{large} models, we also show the predictions of models trained with Twitter data until $2019$, $2020$, $2021$, and $2022$, respectively \cite{loureiro-etal-2022-timelms}.} We first observe that the most widely used \textsc{RoBERTa} \texttt{base} and \texttt{large} models are both outdated in terms of factual knowledge, as they predict the names of PMs that served from $2010$ until $2019$. Next, while the last three models ($2020$-$2022$) answer correctly, the $2019$ model answers the (correct) \textit{first name} of the PM (\texttt{Boris}), not the \textit{surname} (\texttt{Johnson}) which is asked for. 

This is a handy illustration of the many \textit{challenges} in evaluating MLMs for temporal robustness in the LMs-as-KBs framework. First, this $2019$ model would be considered to have made a mistake (as the prediction is different than the gold label and the metric is accuracy), even though the factual knowledge was correct (the name of the PM of the UK). Second, notice that we designed the query to ask for the surname (instead of the name of the PM), as this results in a single mask. The \textsc{lama}-probe and related frameworks do not handle \textit{multi-token} queries for MLMs (e.g., \texttt{Boris Johnson}). Finally, we mark with a \texttt{?} the answers of the first two \textsc{RoBERTa} models, because even though their answers are out-of-date for our current evaluation (October $2022$), their answers could have been correct in an evaluation setting in the time of the training data ($2019$). This illustrates the obscurity of the temporal window in which the model is expected to be correct, if the model is not trained with a temporally-aware design~\cite{lazaridou2021mind, dhingra-etal-2022-time, loureiro-etal-2022-timelms, TemporalWiki}.

In this work, we aim to address such limitations and provide a holistic framework for dynamic benchmarking of masked language models on temporal concept drift, with a focus on facts that change over time. Following the propositions of \citet{kiela-etal-2021-dynabench} and \citet{sogaard-etal-2021-need} that advocate for a focus on \textit{dynamic} (i.e., test sets should not become saturated) and \textit{targeted} (i.e., use of multiple, independent test sets for realistic performance estimates) benchmarking respectively, and building on prior work~\cite{jiang-etal-2020-know, dhingra-etal-2022-time, TemporalWiki}, we create a large open-source test set that can be dynamically updated over time, containing temporal fine-grained subsets of examples that can be used to query masked language models and evaluate their factual knowledge over time. 



\paragraph{Contributions} 
\textbf{(1)} We release \dtemplama{}, an improved version of the static \templama{}~\cite{dhingra-etal-2022-time} test set consisting of Wikidata relations, that is used to evaluate temporal robustness of MLMs. We provide data and code to dynamically keep \dtemplama{} up-to-date over time.\footnote{\url{https://github.com/amazon-science/temporal-robustness}}
\textbf{(2)} We propose a novel evaluation framework to first create temporal splits of test sets of any granularity (month, quarter, year) and then to further create fine-grained splits of facts that are \textit{unchanged}, \textit{updated}, \textit{new} or \textit{deleted}, aiming to improve comprehensiveness (\S\ref{sec:dynamic-templama}). 
\textbf{(3)} We introduce three distinct evaluation views with multiple metrics (\S\ref{sec:evaluation}) to ensure comprehensive results and provide analysis of benchmarking a large set open-source temporal \textsc{RoBERTa} models 
(\S\ref{sec:timelms}).

\section{Related Work}\label{sec:rw}
\paragraph{Temporal Concept Drift}
Evaluation of the robustness of language models on temporal concept drift has seen a rising interest in the recent years. Previous work has focused on methods to continually adapt models over time \cite{AmbaHombaiah2021DynamicLM,Rosin2022TimeMF, lazaridou-internet-augmented}. Another area of research is evaluation of temporal robustness which has been explored both in the upstream LM pretraining task \cite{jiang-etal-2020-know,lazaridou2021mind,dhingra-etal-2022-time, TemporalWiki,loureiro-etal-2022-timelms} and in downstream tasks such as sentiment analysis  \cite{lukes-sogaard-2018-sentiment,Agarwal2022TemporalEO}, named entity recognition \cite{rijhwani-preotiuc-pietro-2020-temporally,onoe-etal-2022-entity}, question answering \cite{TempoQR, streaming-qa}, and rumor detection \cite{yida_eacl2023}. It has also been studied for model explanations~\cite{zhao_emnlp22} and for text classification in legal, biomedical \cite{chalkidis-etal-2022-lw-robust}, and social media \cite{rottger-pierrehumbert-2021-temporal-adaptation} domains.

%
%
\citet{luu-etal-2022-time} explore the setting of temporal misalignment (i.e., training and test data drawn from different periods of time) for both upstream and downstream tasks and find that temporal adaptation should not be seen as a substitute for finding temporally aligned labeled data for fine-tuning.

The closest work to ours is \templama{}~\cite{dhingra-etal-2022-time}. However, we differ across four axes: (i) \templama{} is static, while we provide code to dynamically download facts in a fine-grained fashion from any periods of time (not only yearly), (ii) we evaluate the \textit{same} models over time focusing on the evaluation of robustness over time, we do not explore the best adaptation technique to address the problem, (iii) we do not fine-tune the models to adapt them to the domain/format of the test data, and (iv) we address benchmarking of masked LMs (not auto-regressive) including more evaluation techniques. Finally, similar to our motivation, \citet{TemporalWiki} recently explored lifelong adaptation and evaluation of temporal concept drift in LMs and introduced \textsc{TemporalWiki} for continual adaptation and \textsc{Twiki-Probes} for evaluation. The major difference is that the authors focus on providing corpora to adapt an LM over time, while in our paper we focus on evaluating temporal robustness of LMs. \dtemplama{} is a holistic evaluation framework, while ``\textsc{Twiki-Probes} are not natural sentences; they are factual phrases synthetically generated from a naive concatenation of Subject, Relation, and Object''.

\paragraph{Language Models as Knowledge Bases}
The cloze-style LM evaluation framework for factual knowledge, \textsc{lama}~\citet{petroni-etal-2019-language}, follows the setting depicted in Figure~\ref{fig:intro}. A knowledge base relation is transformed into natural language text with a manually created template and then passed as an input to an LM. The framework is based on treating the output distribution for the \texttt{mask} token as the retrieved answers to the query \cite{https://doi.org/10.48550/arxiv.2204.06031}.
The \textsc{lama} probe has since been extensively used to evaluate factual knowledge in LMs \cite{DBLP:journals/corr/abs-2005-04611,talmor-etal-2020-olmpics, kassner-etal-2021-multilingual, sung-etal-2021-language, dhingra-etal-2022-time, fierro-sogaard-2022-factual}, while other works have been exploring its limitations and ways to improve it \cite{kassner-schutze-2020-negated, haviv-etal-2021-bertese,elazar-etal-2021-measuring,zhong-etal-2021-factual,qin-eisner-2021-learning}. A particular challenge in our experimental setting, is the text compatibility between the model (i.e., its pre-training data) and the format of test examples, named as ``language mismatch'' by \citet{talmor-etal-2020-olmpics}. \citet{dhingra-etal-2022-time} opts to fine-tune the model under evaluation with part of the test set to adapt it to the format of the task. We argue that this process suffers from many caveats; it is inefficient and impractical to fine-tune a model whose capabilities are under evaluation, it risks optimization stability and overfitting issues due to the small training dataset, and enforces extra biases and errors, especially in the case of temporal robustness evaluation. 

\section{Dynamic Benchmarking of Temporal Concept Drift}
In this section we describe in detail the steps to (re)create \dtemplama{}, our dynamically updated test set with facts from Wikidata (\S\ref{sec:dynamic-templama}). We then present the open-source temporal \textsc{RoBERTa} models (\textsc{TimeLMs})~\cite{loureiro-etal-2022-timelms} that we use for benchmarking (\S\ref{sec:timelms}). Finally, we introduce the evaluation framework under which we investigate how well the TimeLMs perform in terms of temporal robustness (\S\ref{sec:evaluation}).  

The research question that we try to address with our work is: \textit{How can we measure temporal drift robustness of PLMs with an evaluation framework that is}: 
\textit{unsupervised} (no labeled downstream data), 
\textit{efficient} (quality test set of facts---no need to run inference on a large corpus to compute perplexity for every token), 
\textit{dynamic} (test set easily generated per request---can be used to dynamically evaluate new concepts over time), 
\textit{general} (option to create test sets of any time granularity), 
and \textit{comprehensive} (battery of targeted test sets that evaluate different LM capabilities and multiple views of evaluation).

\begin{table*}[t]
\small
\centering
\begin{adjustbox}{width=\textwidth}
{\renewcommand{\arraystretch}{1.6}
\setlength{\tabcolsep}{2pt}
\begin{tabular}{lllccc}
\toprule
\textbf{\textsc{Wikidata ID}} & \textbf{\textsc{Relation}  }            & \textbf{\textsc{Template} }                                                                                         
& \textbf{\textsc{\#Facts}}
& \textbf{\textsc{\#Examples}} & \textbf{\textsc{Possible Split(s)}} \\\hline 
P$54$         & member of sports team & \texttt{<subject> plays for <object>.}                      & $3772$           & $50558$        &    $\mathcal{D}^{\textsc{updated}}$   \\
P$69$         & educated at           & \texttt{<subject> attended <object>.}        & $232$                          &    $2420$     &   $\mathcal{D}^{\textsc{updated}}$, $\mathcal{D}^{\textsc{unchanged}}$    \\ 
P$6$          & head of government    & \texttt{<object> is the head of the government of <subject>.} & $578$ &             $7815$     &   $\mathcal{D}^{\textsc{updated}}$    \\ 
P$279$        & subclass of           & \texttt{<subject> is a subclass of <object>.}                 &            $5$ & $70$      &   $\mathcal{D}^{\textsc{new}}$, $\mathcal{D}^{\textsc{updated}}$    \\
\bottomrule
\end{tabular}
}
\end{adjustbox}
\caption{Examples of relations and their corresponding templates that we include in \dtemplama{}. \textsc{\#Facts} denote the unique number of facts for each relation, while \textsc{\#Examples} denotes the total number of example we have collected for each relation in the time range between \texttt{2019-Q1} and \texttt{2022-Q2}. \textsc{Possible Split(s)} indicate the type of fine-grained split that each relation would potentially belong to.}
\label{table:relations}
\end{table*}
\subsection{\textsc{Dynamic-TempLAMA}}\label{sec:dynamic-templama}
\begin{figure}[t!]
    \begin{subfigure}{0.48\textwidth}
        \centering
        \includegraphics[width=\textwidth]{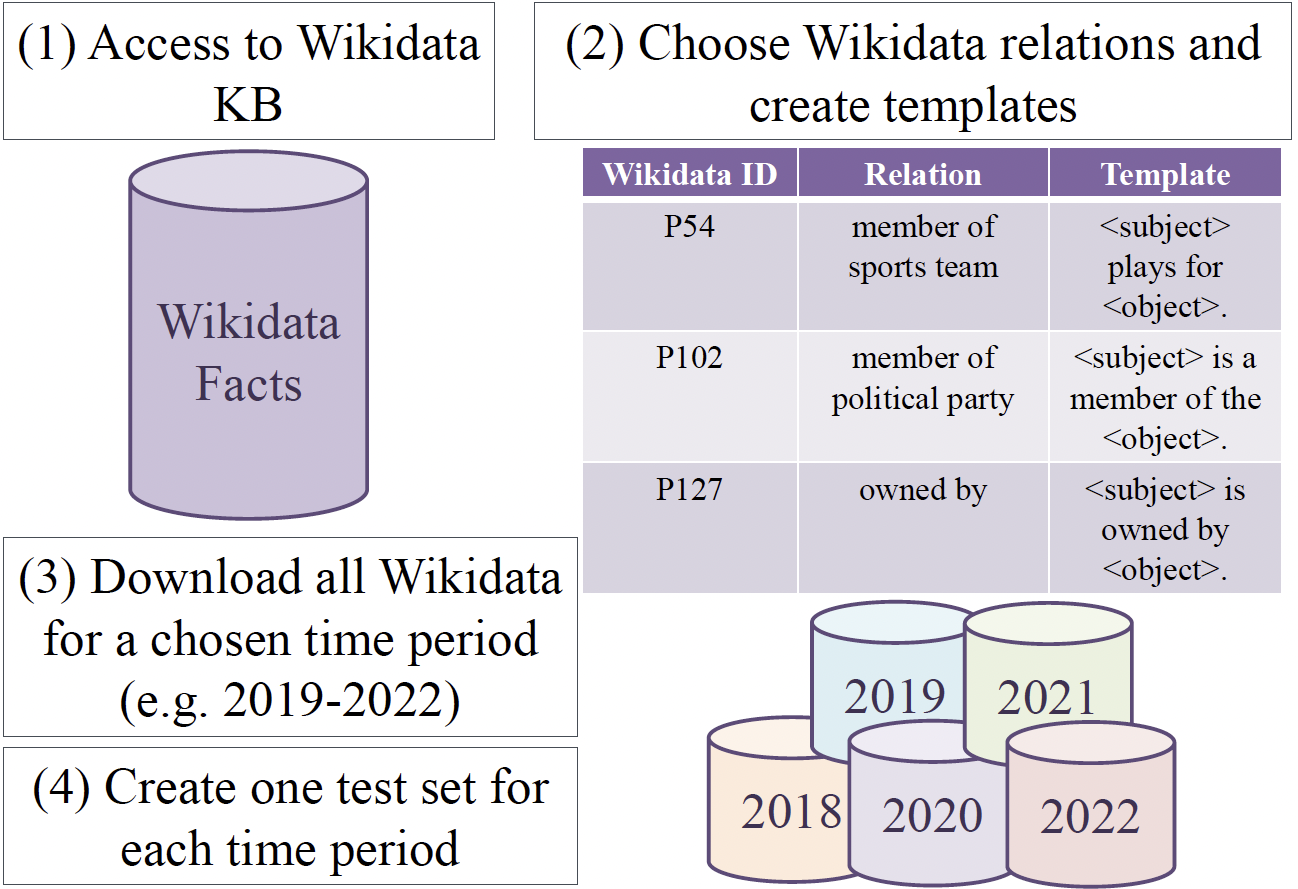}
        \caption{Data collection}\label{fig:data_collection}
    \end{subfigure}%
    \\[\baselineskip]
    \begin{subfigure}{0.48\textwidth}
        \centering
        \includegraphics[width=\textwidth]{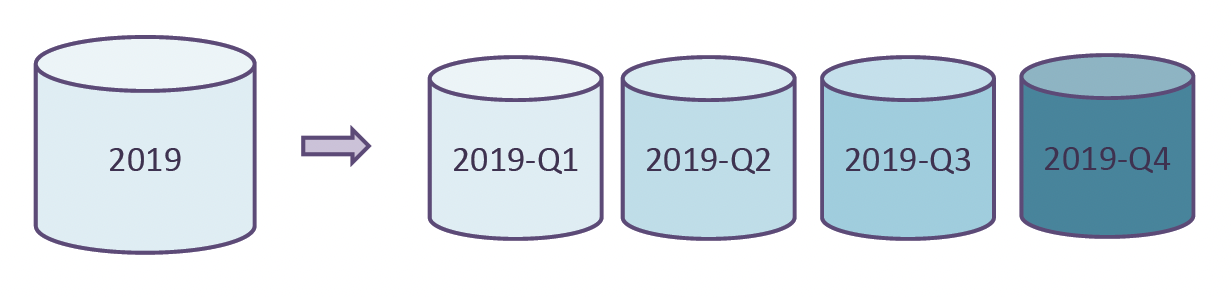}
        \caption{Temporal Splits}\label{fig:temporal_splits}
    \end{subfigure}%
    \\[\baselineskip]
    \begin{subfigure}{0.48\textwidth}
        \centering
        \includegraphics[width=\textwidth]{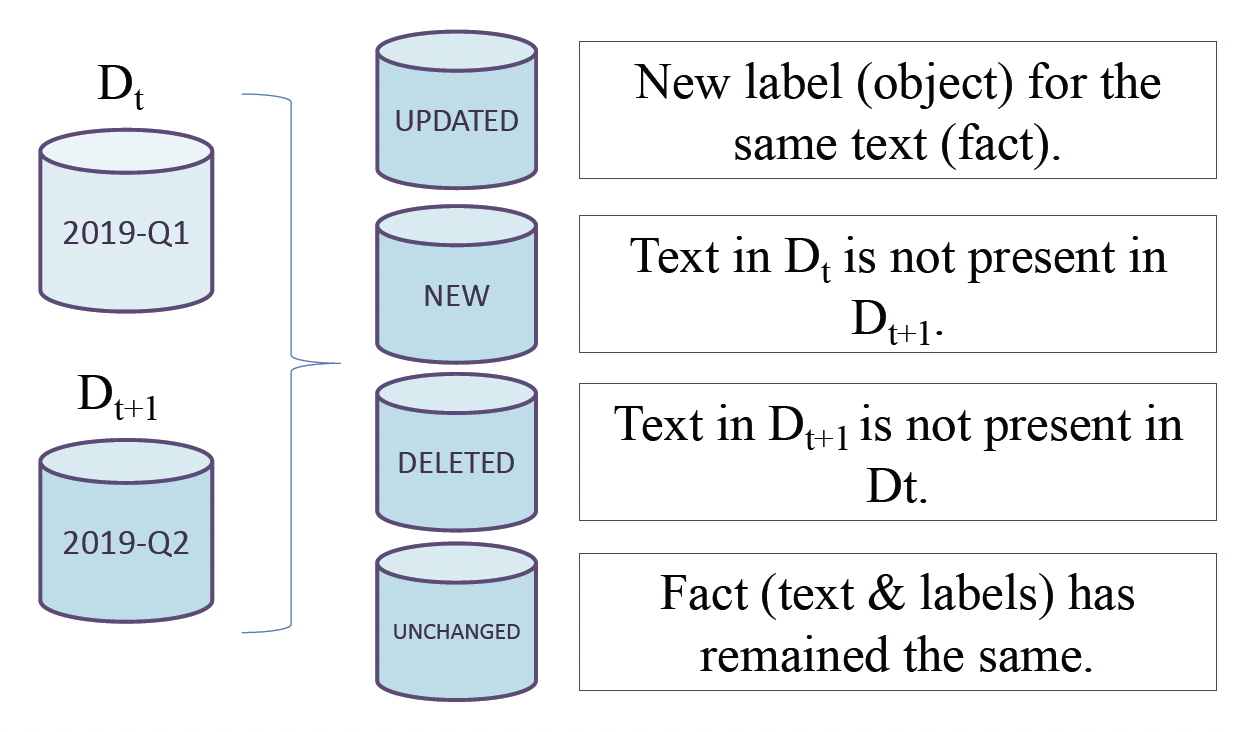}
        \caption{Fine-grained Splits}\label{fig:fingerained_splits}
    \end{subfigure}%
    \caption{The process for creating \dtemplama{}. We first collect data from Wikidata (a), we then divide it to quarterly temporal splits (b) and finally we create more targeted fine-grained sets (c).
    }
    \label{fig:dynamic_templama}
\end{figure}

We base our implementation on the \templama{}~\cite{dhingra-etal-2022-time} code, while we make several changes in terms of accessibility (i.e. option to dynamically update the test set), flexibility (i.e. option to adjust the granularity of the temporal splits) and comprehensiveness (i.e. fine-grained splits and multiple evaluation views). We provide a high-level overview of the process to create \dtemplama{} in Figure~\ref{fig:dynamic_templama}.

\paragraph{Data Collection}
We start the process by selecting a set of \textit{relations} collected from the Wikidata KB (Figure~\ref{fig:data_collection}).\footnote{All possible relations from Wikidata can be found here \url{https://www.wikidata.org/wiki/Wikidata:List_of_properties.}} Specifically, we use the $9$ relations used in the \templama{} dataset, followed by $7$ more that we also decided to collect. We collect all relations from Wikidata in the span of $2019-2022$. We then manually craft a cloze style query, i.e template, for each relation.Table~\ref{table:relations} shows a few examples of relations and templates, along with dataset statistics.\footnote{Details on all relations and templates of \dtemplama{} can be found in Tables~\ref{table:full_relations} \& \ref{table:full_relations_examples} in the Appendix~\ref{sec:data_collection_appendix}.} We explain the data collection process in detail in Appendix~\ref{sec:data_collection_appendix}.

\paragraph{Temporal Splits}
In this stage, we have a very large collection of facts for which we have temporal information (i.e., that the fact is true) in the time range we investigate ($2019-2022$). In the \templama{} dataset, the facts are divided yearly. However, we would ideally like to benchmark temporal models of any time granularity. Specifically, since we benchmark temporal models that are trained quarterly (\S\ref{sec:timelms}), a yearly split would not be useful to evaluate temporal concept drift of the four models trained on each quarter of a year. Consequently, we divide the large set of collected facts per quarter (Figure~\ref{fig:temporal_splits}), while adding the functionality to our implementation to split the facts in any time granularity (monthly, quarterly, yearly). 

\paragraph{Fine-grained Splits}
For a given time range, from timestep $t$ to $t+1$ (e.g. \texttt{2019-Q1}$\rightarrow$\texttt{2019-Q2}), we further create comprehensive test sets that contain examples with \textit{unchanged}, \textit{updated}, \textit{new} or \textit{deleted} facts, denoted by $\Dun, \Dup, \Dnew$ and $\Ddel$ respectively (Figure~\ref{fig:fingerained_splits}). We create these splits to be able to measure different capabilities of the MLM in terms of robustness to temporal concept drift. The motivation for this stems from limitations of prior work~\cite{dhingra-etal-2022-time} to shed light into what kind of data each temporal test set contains. For instance, we pose questions like \textit{How many facts were updated from timestep $t \rightarrow t+1$? How many facts remained unchanged? What was the change? The object or the subject? Are there new facts in timestep $t+1$ that were not present before?} We argue that it is essential to distinguish between these sub-tests, so that each split can target specific capabilities of the LM. First, we can use $\Dun$ to evaluate knowledge preservation (i.e. how well a model can preserve knowledge over time). Second, we can use $\Dup, \Dnew$ and $\Ddel$ to measure adaptation (i.e. how well a model adapts to new information/facts). Finally, we can measure overall temporal robustness by evaluating a temporal model from timestep $t$ on $\Dup$ and $\Dnew$ in timesteps for $t \in [t+1, t+2, ...)$. We believe that this framework is particularly useful for insightful evaluation of methods that aim to adapt language models over time~\cite{10.5555/3524938.3525306, yogatama-etal-2021-adaptive,sun2020lamal,biesialska-etal-2020-continual,jang2022towards, jin-etal-2022-lifelong, chakrabarty2022finetuned}.

\subsection{Temporal Models}\label{sec:timelms}
In contrast with prior work that uses private, in-house models for temporal robustness evaluation that are not accessible by the community~\cite{lazaridou2021mind, dhingra-etal-2022-time}, we instead benchmark a series of open-source temporal models. Despite our aim for transparency, energy efficiency~\cite{DBLP:journals/corr/abs-1906-02243} and reproducibility, we also believe that the \textit{dynamic} nature of the task at hand requires \textit{accessibility} to past, present and future models, to ensure that the findings of evaluation studies in temporal concept drift are meaningful, trustworthy and serve their purpose in evaluating models in a ever-evolving world. Under this assumption, we believe that studies on temporal robustness should ideally build on each other, so that we can have a holistic view as to how these models truly evolve over time.

To this end, we use the Diachronic Language Models (\timelms{}) \cite{loureiro-etal-2022-timelms} that are publicly available in the \texttt{HuggingFace} hub~\cite{huggingface}.\footnote{\url{https://huggingface.co/cardiffnlp}}
\timelms{} are \roberta models~\cite{roberta} trained \textit{quarterly} on Twitter data.
All models are initialised from the original \texttt{roberta-base} model checkpoint and are later trained using data from the previous quarters and the new temporal data from the new time period. For instance, the first model (\texttt{2019-Q4}) was trained with data sampled from Twitter until December $2019$, while the second model (\texttt{2020-Q1}) was trained on the concatenation of all the data used to train \texttt{2019-Q4} and temporally-aligned data sampled from the first quarter of $2020$. There are $11$ \timelms{} in total, from \texttt{2019-Q4} until \texttt{2022-Q2}.

Finally, we would like to draw attention to two specific points. First, all \timelms{} are trained using the same \roberta{} (base) tokenizer and thus have the same vocabulary. This is crucial when evaluating models in a Cloze-style format, like the \textsc{lama}-probe, in order to evaluate fair comparison among the models. Second,
\citet{loureiro-etal-2022-timelms} aim to continue training and releasing \timelms{} every quarter, which is a very important and promising initiative to help with the \textit{dynamic} evaluation of LMs in temporal concept drift in the future.

\subsection{Temporal Concept Drift Evaluation}\label{sec:evaluation}
\paragraph{Single-token probing} Our first evaluation type is single-token probing, which was introduced in the seminal \textsc{lama}-probe work of \citet{petroni-etal-2019-language}. The idea is simple and follows the fill-in-the-blank format. Specifically, we convert each relation using its template to natural language text (see Figure~\ref{fig:dynamic_templama}(a)) replacing the \texttt{<object>} with the mask token (i.e., \texttt{<mask>} for \roberta{}). Then, as shown in Figure~\ref{fig:intro}, we give the prompt as an input to the MLM and obtain a probability distribution over the vocabulary for the \texttt{<mask>} token. We use the metrics from \citet{petroni-etal-2019-language}, that are \texttt{Accuracy}, Mean Reciprocal Rank (\texttt{MRR}) and Precision at k (\texttt{P@k}).\footnote{\texttt{P@k}$=1$, if the gold label is in the top-k predictions of the model, therefore \texttt{P@1} corresponds to \texttt{Accuracy}.} Note that a crucial limitation of this approach is that it considers only facts with single-token objects. This results in trimming down the test sets by $~95\%$, while limiting the actual value of the test (as most facts and concepts contain multiple words).

\paragraph{Multi-token generation}
We aim to address this limitation and include multi-token objects to our evaluation framework. It is important to note that we are benchmarking \textit{masked} language models instead of autoregressive left-to-right language models like \citet{dhingra-etal-2022-time}. This is crucial because the latter, decoder-based family of models,  can be used off-the-shelf to generate multiple tokens. In contrast, MLMs are trained with $~15\%$ of their inputs masked and optimized to predict \textit{only the masked tokens}. We therefore use the formulation introduced by \citet{wang-cho-2019-bert}, that is essentially a decoding-based strategy for MLMs based on Gibbs sampling. Specifically,  we consider the setting that we do not know a priori the correct number of masks for each label. Instead, we enumerate from a single mask up to $M$ masks, i.e., $m=1, ..., M$. Following \citet{jiang-etal-2020-x}, we choose $M=5$, as all our facts are in the English language.
When $m>1$, we add $m$ consecutive masks to the input and we pass the input to the model $m$ times, when each time we sequentially sample each mask from left to right. At each iteration we replace the mask with the corresponding token prediction of the previous iteration. This way, we can extend the \textsc{lama} probe to include multi-token labels in our test set. The setting is entirely different than the single-token approach, as here we have $m$ predictions from the model with an increasing number of tokens, while the correct label can consist of any number of tokens in the range of $1,...,M$. Another difference here is the evaluation metrics. Because we converted the task to text generation, we borrow generation metrics such as \texttt{ROUGE}~\cite{lin-2004-rouge}, while also including standard metrics like \texttt{F$_1$-macro}. Finally, we also include as a metric \texttt{BERT-score}~\cite{Zhang*2020BERTScore:} as an additional informative metric from the perspective of contextual semantics. In effect, we evaluate factual knowledge over time of MLMs, where facts include \textit{multiple correct answers} and each answer consists of \textit{multiple tokens}. We consider a prediction correct if the model correctly predicts \textit{any} of the acceptable answers.

\paragraph{MLM scoring} Finally, as a third lens of evaluation we use the MLM scoring framework of \citet{salazar-etal-2020-masked}. Contrary to the previous approaches, MLM scoring aims to \textit{measure} the probability of the correct answer (i.e., of the masks), instead of \textit{generating} the most probable answer. More specifically, we evaluate MLMs out of the box via their \textit{pseudo-log-likelihood scores} (PLLs), which are computed by masking tokens one by one. PLLs have been widely used to measure the equivalence of perplexity (of autoregressive language models) for MLMs in unlabelled data~\cite{lazaridou2021mind}. Still, computing PLLs for large corpora is a very costly process in terms of time and resources~\cite{loureiro-etal-2022-timelms}. Instead, we propose to combine the \textsc{lama} and MLM scoring frameworks to create an efficient and targeted evaluation framework for temporal factual knowledge.

\begin{table*}[t!]
\small
\centering
\begin{adjustbox}{width=\textwidth}
{\renewcommand{\arraystretch}{1.2}
\setlength{\tabcolsep}{2pt}
\begin{tabular}{lccccccccccccc}
\toprule
\multirow{2}{*}{\textbf{\textsc{Models}}} & \multicolumn{13}{c}{\textbf{\textsc{Temporal Splits}}}                                                                                                         \\
                        & \texttt{2019-Q2} & \texttt{2019-Q3} & \texttt{2019-Q4} & \texttt{2020-Q1} & \texttt{2020-Q2} & \texttt{2020-Q3} & \texttt{2020-Q4} & \texttt{2021-Q1} & \texttt{2021-Q2} & \texttt{2021-Q3} & \texttt{2021-Q4} & \texttt{2022-Q1} & \texttt{2022-Q2}              \\\hline
\texttt{2019-Q4}                 & $34.88$   & $33.96$   & $34.44$   & $34.93$   & $34.76$   & $34.73$   & $34.02$   & $34.18$   & $34.70$    & $34.34$   & $34.92$   & $35.46$   & $35.31$                \\
\texttt{2020-Q1}                 & $24.47$   & $24.01$   & $24.45$   & $24.67$   & $24.59$   & $24.44$   & $23.98$   & $23.94$   & $24.25$   & $23.96$   & $24.20$    & $24.5$    & $24.42$                \\
\texttt{2020-Q2}                 & $22.94$   & $22.29$   & $22.92$   & $23.24 $  & $23.23 $  & $23.12$   & $22.57$   & $22.55$   & $22.90$    & $22.59$   & $22.91$   & $23.23$   & $23.11$                \\
\texttt{2020-Q3}                 & $22.39$   & $21.87$   & $22.22$   & $22.60 $  & $22.52$  & $22.42$   & $21.99$   & $22.00$   & $22.29$    & $21.92$   & $22.18$   & $22.42$   & $22.30$                \\
\texttt{2020-Q4}                 & $25.56$ & $25.28$ & $25.68$ & $25.96$ & $25.89$ & $25.79$ & $25.51$ & $25.44$ & $25.71$ & $25.50$  & $25.69$ & $25.97$ & $25.72$ \\
\texttt{2021-Q1}                 & $25.76$ & $25.28$ & $25.91$ & $26.18$ & $26.14$ & $26.18$ & $25.75$ & $25.63$ & $25.99$ & $25.77$ & $26.01$ & $26.32$ & $26.02$ \\
\texttt{2021-Q2}                 & $23.75$ & $23.47$ & $23.94$ & $24.10$  & $24.10$  & $24.12$ & $23.63$ & $23.60$  & $24.05$ & $23.75$ & $24.12$ & $24.37$ & $24.16$ \\
\texttt{2021-Q3}                 & $22.95$ & $22.61$ & $23.00$  & $23.14$ & $23.12$ & $23.16$ & $22.84$ & $22.77$ & $23.00$  & $22.82$ & $23.03$ & $23.30$  & $23.06$ \\
\texttt{2021-Q4}                 & $23.37$ & $23.01$ & $23.41$ & $23.59$ & $23.55$ & $23.68$ & $23.37$ & $23.27$ & $23.60$  & $23.40$  & $23.58$ & $23.76$ & $23.61$ \\
\texttt{2022-Q1}                 & $24.25$ & $23.83$ & $24.42$ & $24.56$ & $24.57$ & $24.68$ & $24.40$  & $24.26$ & $24.52$ & $24.35$ & $24.51$ & $24.71$ & $24.58$ \\
\texttt{2022-Q2}                 & $\textbf{21.48}$ & $\textbf{20.95}$ & $\textbf{21.42}$ & $\textbf{21.59}$ & $\textbf{21.57}$ & $\textbf{21.61}$ & $\textbf{21.25}$ & $\textbf{21.12}$ & $\textbf{21.44}$ & $\textbf{21.13}$ & $\textbf{21.31}$ & $\textbf{21.49}$ & $\textbf{21.39}$\\
                        \bottomrule
\end{tabular}
}
\end{adjustbox}
\caption{MLM scoring (median pseudo-log-likelihood scores) averaged for each temporal split.}\label{table:overall_pppl}
\end{table*}

\subsection{Dataset Analysis}\label{sec:dataset_analysis}
We consider different subsets of the \dtemplama{} test sets for the three different evaluation settings (\S\ref{sec:evaluation}). For the multi-token and MLM scoring settings, we keep the full dataset, for single-token we first tokenize the labels 
and keep only the test examples that have at least one label with a single token. This results in a very aggressive filtering of the dataset. Specifically, each quarterly temporal split consists of $~8500$ test examples on average for the multi-token setting, but for the single-token this results in only $~450$ examples, marking a loss of $95\%$ of the data.\footnote{Table~\ref{table:dtemplama_stats} in the Appendix shows all the statistics in detail.} Additionally, the distribution of the fine-grained splits is of great interest, as it will shape the interpretation of the results and the general challenges of the evaluation framework. $\mathcal{D}^{\textsc{updated}}$ and $\mathcal{D}^{\textsc{unchanged}}$ (i.e., the splits of the most interest) constitute around $96\%$ and $0.3\%$, respectively, of the total examples for the single-token evaluation, and $95\%$ and $1.8\%$ for the multi-token. This is arguably a very skewed distribution, showing the importance of our work in diving the temporal splits into further fine-grained splits. This is essential, because we would have different expectations for a model trained on timestep $t$ while tested on data from both $t$ and $t-1$; for \textit{unchanged} facts it would be desirable to keep equal performance in both sets (i.e., knowledge preservation \S\ref{sec:knowledge_preservation}), while for updated facts we would like to see improved performance in timestep $t$ (i.e., adaptation \S\ref{sec:adaptation}).
\begin{figure}[!t]
\centering
    \begin{subfigure}{0.5\columnwidth}
        \centering
        \includegraphics[width=\textwidth]{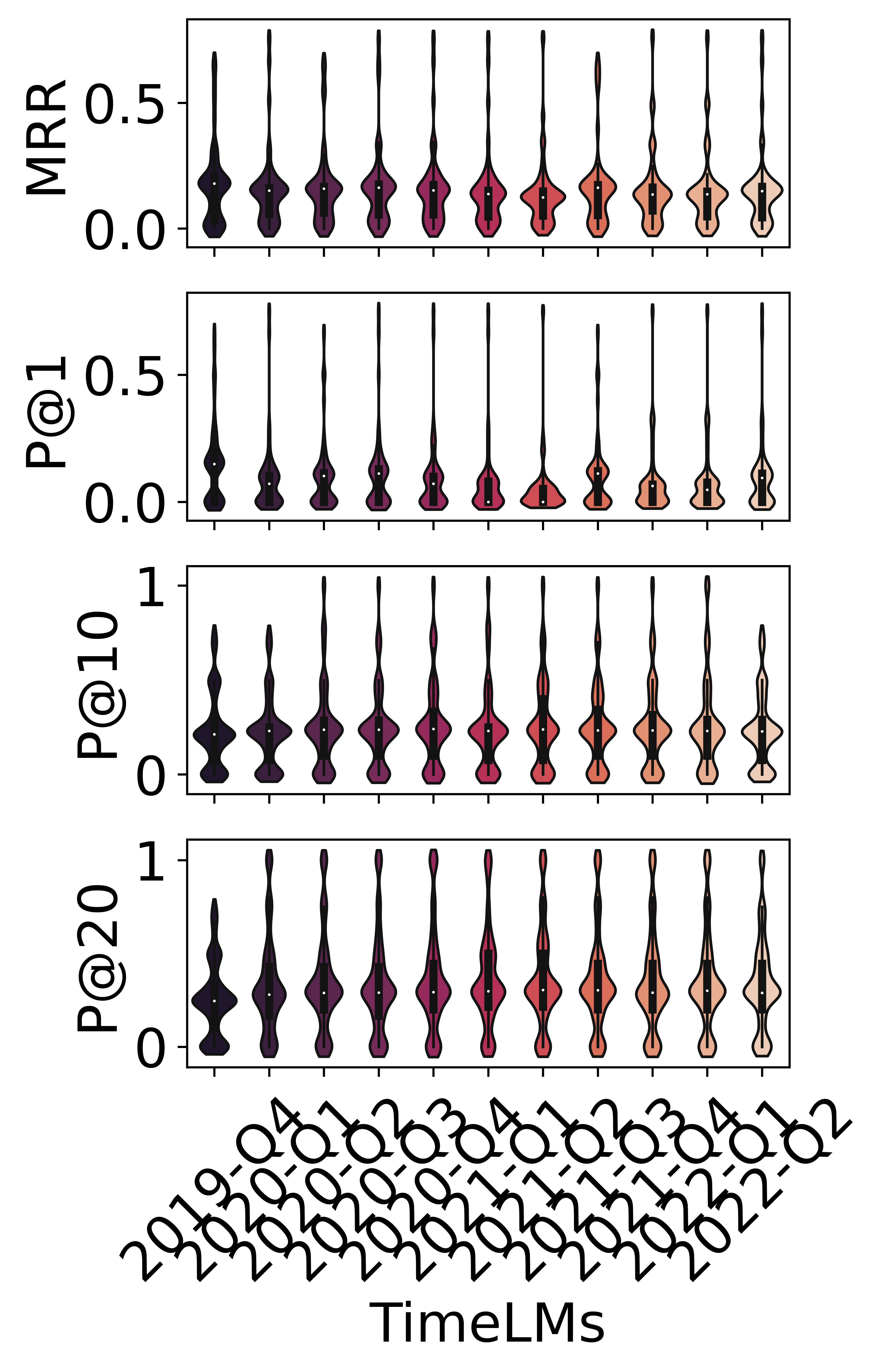}
        \caption{Single-token}\label{fig:single_overall}
    \end{subfigure}%
    \begin{subfigure}{0.5\columnwidth}
        \centering
        \includegraphics[width=\textwidth]{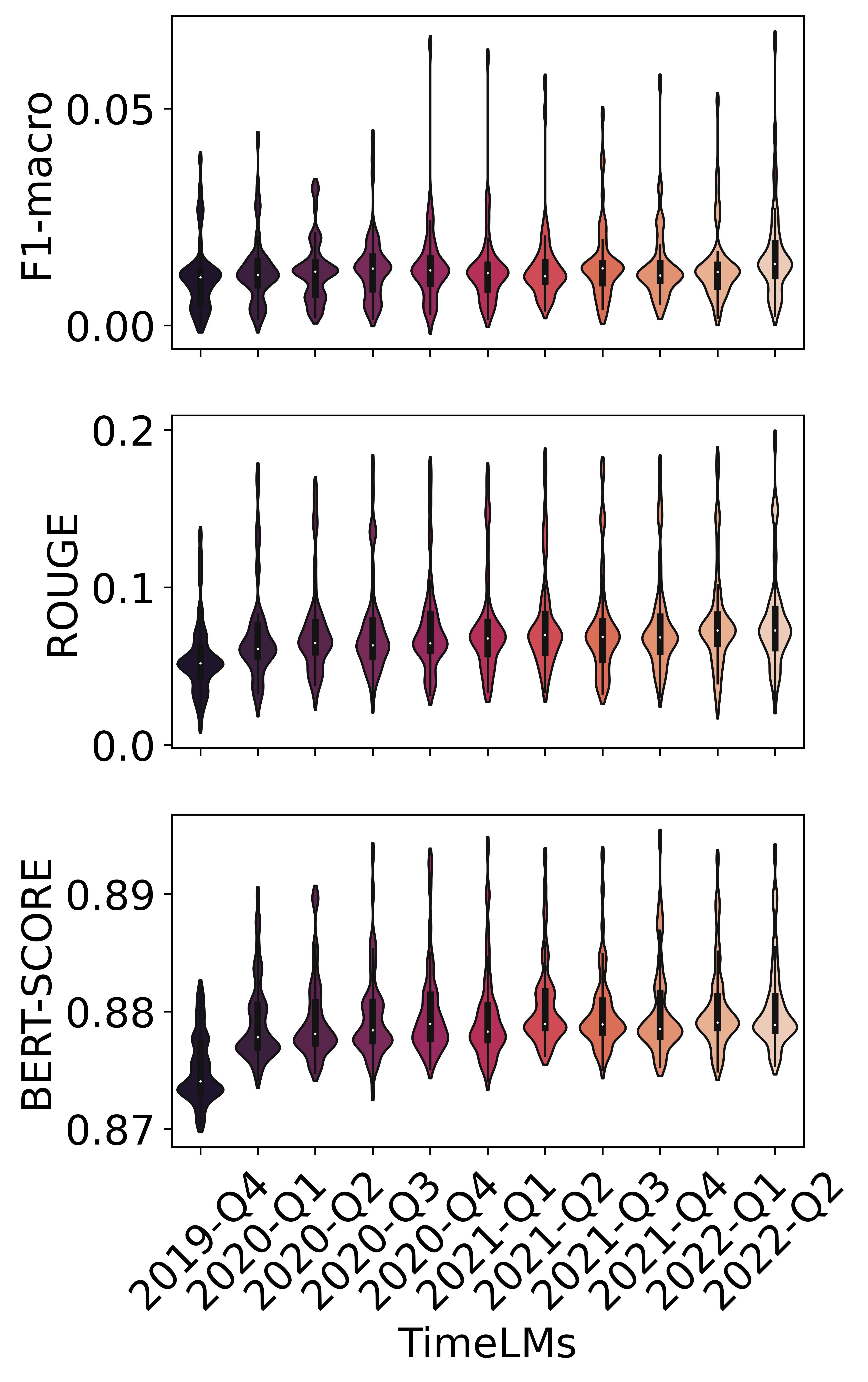}
        \caption{Multi-token}\label{fig:multi_overall}
    \end{subfigure}%
    \caption{\textit{Overall} performance over time ($2019-2022$) for both single and multi-token evaluation. $X$-axis corresponds to the \timelms{} and the $Y$-axis to different metrics depending on the type of the evaluation.
    }
    \label{fig:overall_performance}
\end{figure}
\section{Results}

\begin{figure*}[!t]
\centering
    \begin{subfigure}{\columnwidth}
        \centering
        \includegraphics[width=\textwidth]{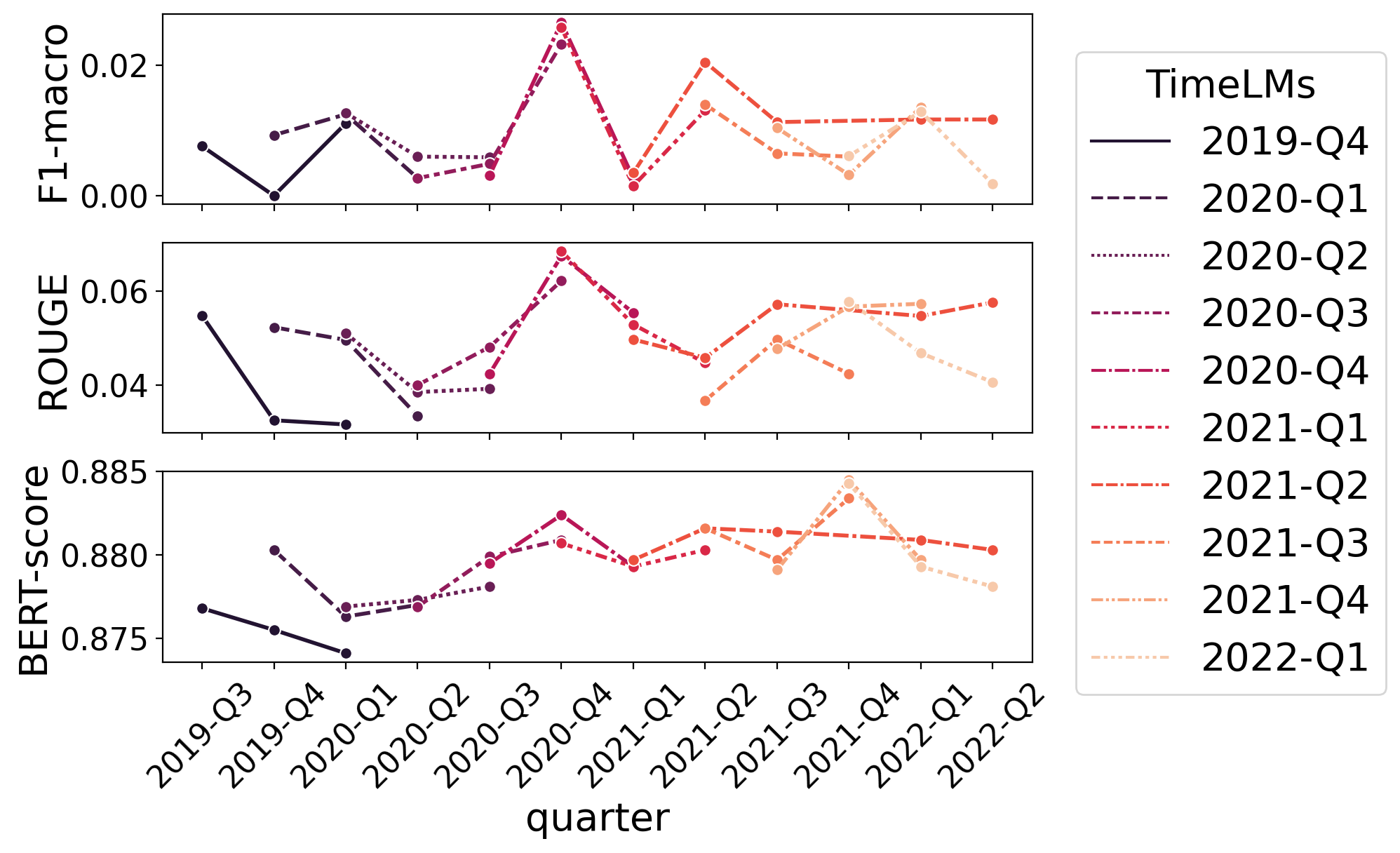}
        \caption{\textsc{Updated} split.}
    \end{subfigure}%
    \begin{subfigure}{\columnwidth}
        \centering
        \includegraphics[width=\textwidth]{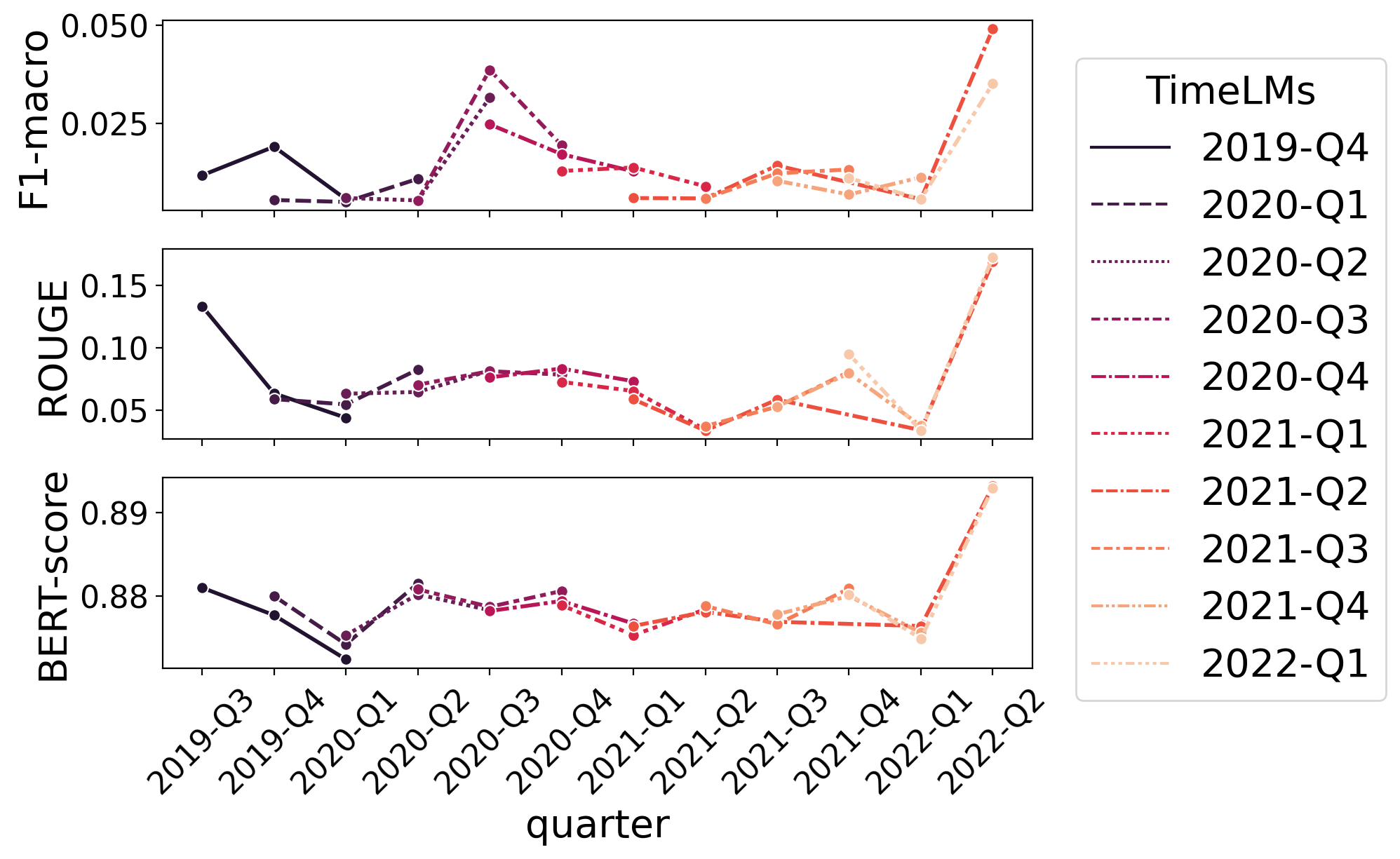}
        \caption{\textsc{New} split.}
    \end{subfigure}%
    \caption{Multi-token evaluation for evolving and emerging facts.
    }
    \label{fig:updated_new}
\end{figure*}

\subsection{Temporal robustness}\label{sec:temporal_robustness}
We first evaluate \textit{temporal robustness} of the $11$ \timelms{}, defined as the \textit{overall} performance over time (\S\ref{sec:dynamic-templama}). Figure \ref{fig:overall_performance} shows the average performance in \textit{all} temporal and fine-grained splits in the time range from \texttt{2019-Q4} to \texttt{2022-Q2} for two types of evaluation, single-token probing and multi-token generation. For the former evaluation type,
(Fig.~\ref{fig:single_overall}), 
all models perform similarly for all metrics. However, when we evaluate multi-token generation the models gradually improve over time.
(Fig.~\ref{fig:multi_overall}). 
This difference shows the importance of considering \textit{multiple views} and evaluations for the same LM capability (i.e., temporal robustness). 

\begin{figure}[t!]
\centering
    \begin{subfigure}{0.8\columnwidth}
        \centering
        \includegraphics[width=\textwidth]{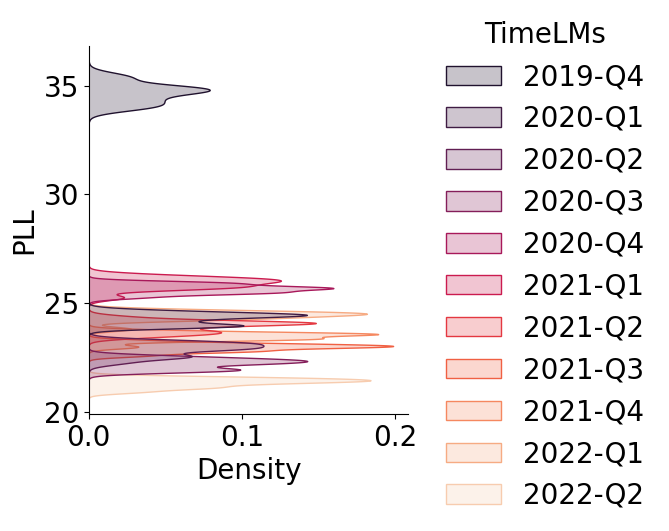}
    \end{subfigure}%
    \caption{\textit{Overall} PLL distributions for \timelms{}.
    }
    \label{fig:overall_pll_density}
\end{figure}

We attribute the similar single-token performance to the fact that these temporal datasets contain almost exclusively \textit{unchanged} facts (\S\ref{sec:dataset_analysis}). It is therefore a positive outcome to observe that \timelms{} can preserve acquired knowledge (\S\ref{sec:knowledge_preservation}). The findings for overall multi-token evaluation corroborate the intuition that more recent models, that are trained with temporal data of the entire range, should perform better than ``past'' (e.g. \texttt{2020}) models that have not seen ``future'' data (e.g. \texttt{2022})
during training. 
We also provide the overall results with MLM scoring in Table~\ref{table:overall_pppl}. We also observe that the last model performs best across all temporal splits, showing the effectiveness of adaptation with more recent unlabelled data  (\S\ref{sec:timelms}). Even though we observe that this pattern holds for most temporal splits (i.e., scores improving for each column $\downarrow$), the \texttt{2020-Q4} and \texttt{2021-Q1} \timelms{} produce worse PLL scores than their previous or later versions. This is more evident in the overall density plot in Figure~\ref{fig:overall_pll_density}. This finding entails that either the distribution shift in these quarters was a lot stronger than the other temporal periods, or the training of these particular models was not as successful as it would have been expected.

\begin{table*}[h]
\small
\centering
\begin{adjustbox}{width=\textwidth}
{\renewcommand{\arraystretch}{1.2}
\setlength{\tabcolsep}{2pt}
\begin{tabular}{lllccc}
\toprule
& \textsc{Example Input }  & \textsc{Ground Truth Labels}  &\textsc{\#Tokens} & \textsc{\#Answers} & \textsc{Split}   \\ \hline
\multirow{2}{*}{$1$}  & \multirow{2}{*}{\texttt{Alex Morgan plays for \_X\_.}} & \texttt{United States women's national soccer team} & $7$     & \multirow{2}{*}{$2$}         & \multirow{2}{*}{\texttt{2021-Q4}} \\
  &  &  \texttt{Orlando Pride} &   $2$  &         &  \\\hline
%
 \multirow{3}{*}{$2$}  &  \multirow{3}{*}{\texttt{Cristiano Ronaldo plays for \_X\_.}}  &   \texttt{Juventus F.C.}         &    $5$      &    $1$       &   \texttt{2021-Q2}      \\
 &  &   \texttt{Juventus F.C., Manchester United F.C.}         &   $5,6$       &     $2$      &    \texttt{2021-Q3}     \\
 & &     \texttt{Manchester United F.C.}       &      $6$    &  $1$         &     \texttt{2021-Q4}    \\\hline
 \multirow{3}{*}{$2$}  &  \multirow{3}{*}{\texttt{\_X\_ is the head of the government of Italy.}} &    \texttt{Giuseppe Conte}  &    $5$        &    $1$      &   \texttt{2020-Q4}      \\
  &   &  \texttt{Giuseppe Conte, Mario Draghi}   &   $5,3$         &    $2$      &   \texttt{2021-Q1}   \\
  &   &  \texttt{Mario Draghi}  &   $3$         &    $1$      &   \texttt{2021-Q2}   \\
  \bottomrule
\end{tabular}
}
\end{adjustbox}
\caption{Qualitative analysis of certain examples in \dtemplama{}.}\label{table:qual_examples}
\end{table*}

\subsection{Knowledge preservation}\label{sec:knowledge_preservation}
\begin{figure}[t!]
\centering
    \begin{subfigure}{\columnwidth}
        \centering
        \includegraphics[width=\textwidth]{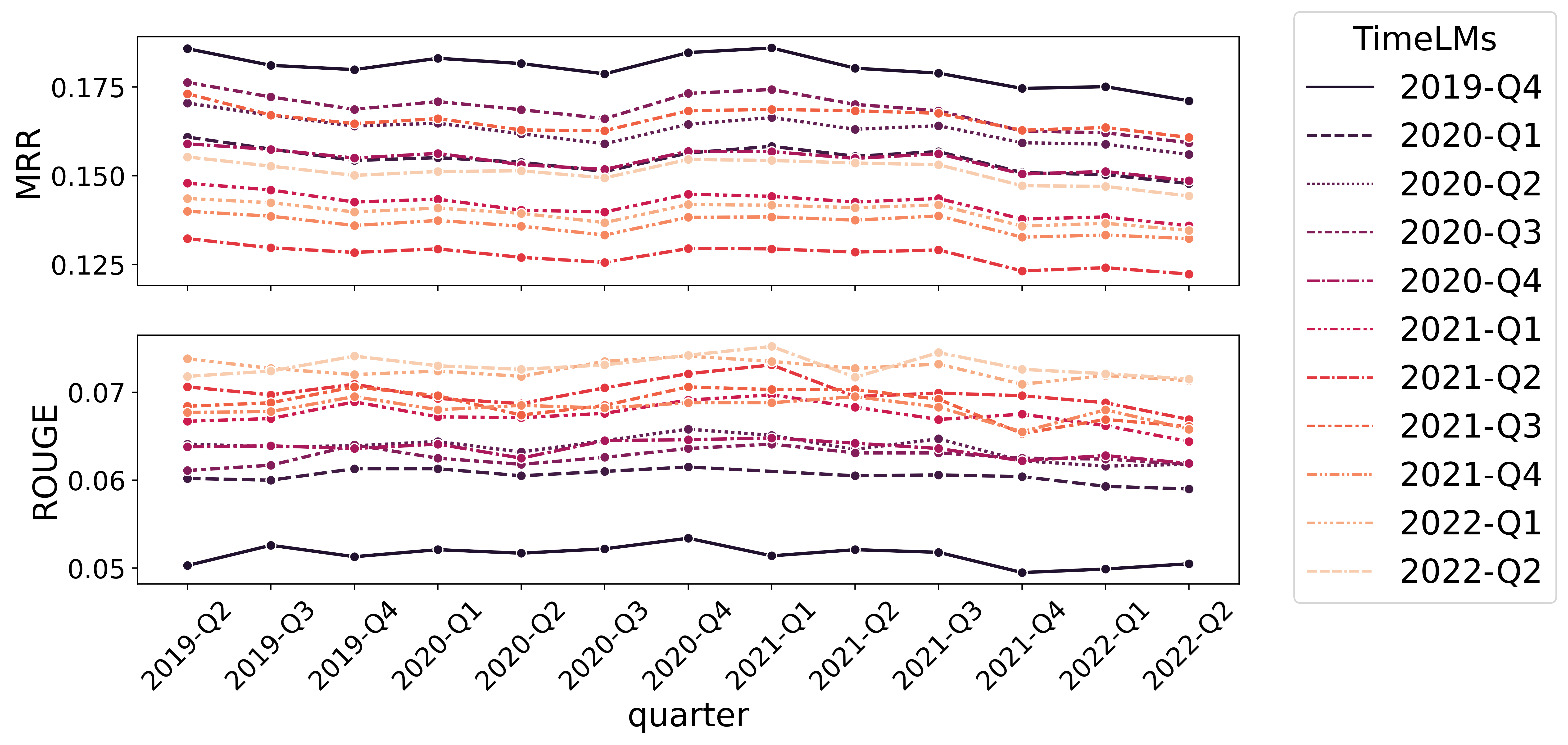}
    \end{subfigure}%
    \caption{Single and multi-token evaluation for the \textsc{Unchanged} split.
    }
    \label{fig:unchanged}
\end{figure}

We use the $\mathcal{D}^{\textsc{unchanged}}$ split to evaluate the capability of MLMs to preserve knowledge over time. Figure~\ref{fig:unchanged} shows that for both single and multi-token evaluation all \timelms{} demonstrate similar performance over time, showing strong knowledge preserving skills. Surprisingly, different metrics show different patterns among the models for a single split. While in general we should not compare the performance of the single model over time (as the test sets are different), the comparision is valid in this case because the splits contain unchanged facts, and hence most temporal test sets are almost identical. All plots are shown in Figure~\ref{fig:unchanged_full} in the Appendix.

\subsection{Adaptation to emerging \& evolving concepts}\label{sec:adaptation}
Finally, we use the $\mathcal{D}^{\textsc{new}}$ and $\mathcal{D}^{\textsc{updated}}$ splits for evaluation of emerging and evolving concepts, respectively. Here to ensure fair comparison, we evaluate the \timelms{} for a specific time window; for each model trained on 
timestep 
$t$, we keep the 
test 
sets from $t-1$, $t$ and $t+1$. We observe in Figure~\ref{fig:updated_new} that 
in these cases 
the results vary among the models. There is not a very clear pattern as before, so case-by-case examination would be required. Still, a common pattern for the \textsc{Updated} split is that the middle set tends to have the highest performance ($\wedge$ shape). This means that models manage to effectively adapt to the updated facts of that timestep ($t$), but on the next timestep ($t+1$) they underpeform as they are unaware of the factual changes, thus requiring adaptation. We provide all plots in the Appendix, including the \textsc{Deleted} split, which is more difficult to interpret intuitively (i.e., why are some facts deleted from Wikidata after a certain point?).

\section{Qualitative Analysis}\label{sec:qual_analysis}
Table~\ref{table:qual_examples} provides some examples from the \dtemplama{} test set that can help us further interpret our results and inspect existing challenges. We first observe that all examples have multi-token labels (i.e., objects from the \texttt{Subject-relation-object} format) and are in effect discarded in the single-token evaluation setup, making the inclusion of multiple views essential for this task. 

More specifically, in $1$, we observe that one label (\texttt{United States women's national soccer team}) has more than $M=5$ tokens. It is therefore excluded even from the multi-token the test set, leaving MLM scoring to be the only method that could evaluate it. Interestingly, we manually tested the \texttt{2021-Q4} temporal model and found that it produces $1.6$ and $307.3$ average PLL scores for the two options respectively, making the disregarded label a far more confident prediction. 

In the second and third example, we observe how the correct answer for the query changes over time, making the granularity of the evaluation (i.e., yearly, quarterly, monthly) an important factor in the correct assessment of the model's temporal factual knowledge. For instance, for the example $3$, we can carefully inspect how the predictions of the models change for facts that change over time (Table~\ref{table:qual_predictions}). However, even though PLL scores can follow intuitive temporal patterns (i.e., the PLL value can increase or decrease according to the point in time that the fact has changed), comparison between scores is not always helpful (i.e., word frequency can obscure factual knowledge) leaving room for improving the \textsc{lama} formulation.







\begin{table}[t]
\small
\centering
\begin{adjustbox}{width=0.85\columnwidth}
{\renewcommand{\arraystretch}{1.2}
\setlength{\tabcolsep}{2pt}
\begin{tabular}{lcc}
\toprule
\textsc{TimeLms} & \texttt{Guiseppe Conte}  & \texttt{Mario Draghi} \\ \hline
\texttt{2020-Q4} & $3.8$ & $33.3$ \\
\texttt{2021-Q1} & $3.5$ & $22.7$ \\
\texttt{2021-Q2} & $3.5$ & $25.7$ \\
\texttt{2021-Q3} & $3.8$ & $23.8$ \\
  \bottomrule
\end{tabular}
}
\end{adjustbox}
\caption{PLL scores for Example $2$ from Table~\ref{table:qual_examples}.}\label{table:qual_predictions}

\end{table}

\section{Conclusion \& Future Work}
We addressed MLMs' robustness on temporal concept drift and introduced \dtemplama{}: a dataset for dynamic benchmarking of factual knowledge in temporal, fine-grained splits,
from \texttt{2019-Q4} to \texttt{2022-Q2} 
that contain facts over time. 
We release our codebase to \textit{dynamically} update the 
current 
test set over time and the option to extend it with custom (i) templates, (ii) relations from Wikidata, (iii) any period of time (years) and (iv) granularity of time (month/quarter/year). We include \textit{multiple views of evaluation}, showing that it is essential in order to properly interpret the results of our benchmarking study of $11$ temporal \roberta{} models. We consider experimentation with improving MLM decoding and addressing ``domain mismatch'' as open areas of research for future work. Our code can be found at \url{https://github.com/amazon-science/temporal-robustness}.

\section*{Acknowledgements}
We would like to thank the anonymous reviewers and the AWS AI team for their valuable feedback on the paper. We also thank Robert Vacareanu and Karthikeyan K for their help with running experiments. 

\section*{Limitations}
\paragraph{Lower bound estimate} A very common issue with the \textsc{lama} probe evaluation framework \cite{petroni-etal-2019-language} is that it constitutes a lower bound estimate for its performance on factual knowledge retrieval. Specifically, if a model performs well, one can infer that it has the tested reasoning skill. However, failure does not entail that the reasoning skill is missing, as it is possible that there is a problem with the lexical-syntactic construction we picked~\cite{talmor-etal-2020-olmpics}. Any given prompt only provides a lower bound estimate of the knowledge contained in an LM \cite{jiang-etal-2020-know}.
\paragraph{Domain mismatch} Despite the advantages of zero-shot evaluation, performance of a model might be adversely affected by mismatches between the language the pre-trained LM was trained on and the language of the examples in our tasks \cite{jiang-etal-2020-know}. It is quite possible that a fact that the LM does know cannot be retrieved due to the prompts not being effective queries for the fact \cite{jiang-etal-2020-know}. Prior work proposes to fine-tune the model with a small set of examples taken from the test set (and removed of course) in order to address the incompatibility problem or ‘language mismatch’ \cite{talmor-etal-2020-olmpics, dhingra-etal-2022-time}. We argue that this process suffers for multiple limitations, such as that it not practical for a fast evaluation of the capabilities of a PLM at hand and it faces optimization stability issues due to the small training dataset, inter alia. The major limitation, however, is that such fine-tuning enforces extra biases and errors, especially in the case of temporal robustness evaluation.
\paragraph{MLM decoding (multi-token labels)} In this work we tried to address the problem of \textit{decoding from masked language models}, by incorporating two distinct approaches to the evaluation framework; multi-token generation with MLMs~\cite{wang-cho-2019-bert} and MLM scoring~\cite{salazar-etal-2020-masked}. Still, we observe that both methods provide results that are hard to interpret (\S\ref{sec:qual_analysis}), leaving the problems of (i) decoding or generating multiple tokens from MLMs and (ii) evaluation of factual knowledge in LMs as open areas of research.
\paragraph{Manual Templates} For \textsc{lama}-style probing \cite{petroni-etal-2019-language}, prior work creates the templates \textit{manually}. This is a limitation both in terms of scale (i.e., generalization to many different kinds of inputs) and consistency (i.e., how do models perform with minimal changes to their inputs?). LMs do not reason in an abstract manner and are context-dependent~\cite{talmor-etal-2020-olmpics}. It is therefore essential to address this problem and include functionalities to incorporate a set of diverse templates for each evaluation setup.
\paragraph{English Twitter MLMs} Finally, our dataset, \dtemplama{}, following prior work \cite{dhingra-etal-2022-time}, collects and evaluates facts from the Wikidata in the \textit{English} language alone, and benchmarks RoBERTa language models trained in English Twitter data. We understand that this is a limitation and further data collection and experimentation in more languages would be strongly encouraged.


\bibliography{custom}
\bibliographystyle{acl_natbib}

\clearpage 
\appendix

\section{Appendix}\label{sec:appendix}

\subsection{Data Collection for \dtemplama{}}\label{sec:data_collection_appendix}
Following ~\citet{dhingra-etal-2022-time}, we identify all facts in the Wikidata snapshot,
which have either a start or an end date after $2010$ and whose subjects and objects are both entities with Wikipedia pages.1 Among these $482$K facts, we identify subject and relation pairs which have multiple objects at different times and select $16$ relations with the most such subjects. 
Then, for these relations we manually write template cloze queries (i.e., templates) and populate them with the
$1000$ most frequent subjects per relation. For each
subject and each relation we gather all the objects
with their associated time interval and construct a
separate query for each year in that interval. When
intervals for the object entities overlap, we add all
of them to the list of correct answers. The query
and the corresponding year form the input texts  and
the temporal information $t$, while the object entity is the target that we want to predict (i.e., gold label). In contrast to \citet{dhingra-etal-2022-time}, we do extra temporal divisions. Specifically, we get each yearly split and divide it further in quarterly splits (\S\ref{sec:dynamic-templama}, Figure~\ref{fig:temporal_splits}), following the same algorithm.

\begin{table*}[t]
\centering
\begin{adjustbox}{width=\textwidth}
{\renewcommand{\arraystretch}{1.2}
\setlength{\tabcolsep}{2pt}
\begin{tabular}{cccccc|cc|c}
\toprule
\textbf{\textsc{Temporal Split}} & \textbf{\textsc{Unchanged} } & \textbf{\textsc{Updated}} & \textbf{\textsc{Deleted}} & \textbf{\textsc{New}} & \textbf{\textsc{Total} }& \textbf{\textsc{\%Unchanged}} & \textbf{\textsc{\%Updated} }&\textbf{\textsc{\%Lost}}\\ \hline
\texttt{2019-Q2}   & $ 479|8523 $ & $ 1|165 $  & $ 7|124 $  & $ 9|121 $  & $ 496|8933 $ & $ 96.6|95.4 \%$ & $ 0.2|1.8\%$  & $94.4\%$\\
\texttt{2019-Q3}   & $ 451|8154 $ & $ 3|248 $  & $ 36|430 $ & $ 5|205 $  & $ 495|9037 $ & $ 91.1|90.2 \%$ & $ 0.6|2.7 \%$ & $94.5\%$\\
\texttt{2019-Q4}   & $ 454|8271 $ & $ 0|151 $  & $ 3|140 $  & $ 12|120 $ & $ 469|8682 $ & $ 96.8|95.3 \%$ & $ 0.0|1.7 \%$ & $94.6\%$\\
\texttt{2020-Q1}   & $ 456|8243 $ & $ 3|296 $  & $ 9|126 $  & $ 15|273 $ & $ 483|8938 $ & $ 94.4|92.2 \%$ & $ 0.6|3.3 \%$ & $94.6\%$\\
\texttt{2020-Q2}   & $ 470|8451 $ & $ 0|92 $   & $ 2|95 $.  & $ 2|59 $   & $ 474|8697 $ & $ 99.2|97.2 \%$ & $ 0.0|1.1 \%$ & $94.5\%$\\
\texttt{2020-Q3}   & $ 446|8254 $ & $ 2|179 $  & $ 26|238 $ & $ 10|133 $ & $ 484|8804 $ & $ 92.1|93.8 \%$ & $ 0.4|2.0 \%$ & $94.5\%$\\
\texttt{2020-Q4}   & $ 452|8298 $ & $ 2|124 $  & $ 4|111 $  & $ 5|97 $   & $ 463|8630 $ & $ 97.6|96.2 \%$ & $ 0.4|1.4 \%$ & $94.6\%$\\
\texttt{2021-Q1}   & $ 453|8238 $ & $ 1|269 $  & $ 4|131 $  & $ 14|215 $ & $ 472|8853 $ & $ 96.0|93.1 \%$ & $ 0.2|3.0 \%$ & $94.7\%$\\
\texttt{2021-Q2}   & $ 460|8344 $ & $ 2|90 $   & $ 7|128 $  & $ 5|76 $   & $ 474|8638 $ & $ 97.0|96.6 \%$ & $ 0.4|1.0 \%$ & $94.5\%$\\
\texttt{2021-Q3}   & $ 445|8164 $ & $ 2|164 $  & $ 19|220 $ & $ 2|99 $   & $ 468|8647 $ & $ 95.1|94.4 \%$ & $ 0.4|1.9 \%$ & $94.6\%$\\
\texttt{2021-Q4}   & $ 443|8213 $ & $ 1|128 $  & $ 4|82 $   & $ 5|90 $   & $ 453|8513 $ & $ 97.8|96.5 \%$ & $ 0.2|1.5 \%$ & $94.7\%$\\
\texttt{2022-Q1}   & $ 442|8189 $ & $ 1|111 $  & $ 7|117 $  & $ 6|126 $  & $ 456|8543 $ & $ 96.9|95.9 \%$ & $ 0.2|1.3 \%$ & $94.7\%$\\
\texttt{2022-Q2}   & $ 446|8287 $ & $ 0|56 $   & $ 2|40 $   & $ 2|34 $   & $ 450|8417 $ & $ 99.1|98.5 \%$ & $ 0.0|0.7 \%$ & $94.7\%$\\
   \bottomrule
\end{tabular}
}
\end{adjustbox}
\caption{Total number of examples for each temporal and fine-grained split in \dtemplama{}. We show both the \textit{single-token} and the \textit{multi-token} datasets (up to $M=5$ tokens). Cell scheme to be read \textit{single} | \textit{multi}. \textsc{\%Unchanged} and \textsc{\%Updated} show the percentage of the total examples that are part of the \textsc{Unchanged} and \textsc{Updated} set respectively. \textsc{\%Lost} shows the percentage of examples we lose when we filter out the dataset for the \textit{single-token} evaluation setting.}
\label{table:dtemplama_stats}
\end{table*}

\begin{table*}[t]
\small
\centering
\resizebox{0.95\textwidth}{!}{
{\renewcommand{\arraystretch}{1.6}
\begin{tabularx}
{\linewidth}{
    >{\hsize=0.5\hsize}X
    >{\hsize=1.1\hsize}X
    >{\hsize=2.7\hsize}X
    >{\hsize=0.5\hsize}X
    >{\hsize=0.7\hsize}X
    >{\hsize=0.5\hsize}X
  }
\toprule
\textbf{\textsc{Wikidata ID}} & \textbf{\textsc{Relation} }             & \textbf{\textsc{Template} }                                                                                         
& \textbf{\textsc{\#Facts}}
& \textbf{\textsc{\#Examples}} & \textbf{\textsc{Possible Split(s)} }\\\hline
P$54$         & member of sports team & \texttt{<subject> plays for <object>.}                      & $3772$           & $50558$        &    $\mathcal{D}^{\textsc{updated}}$   \\\hline
P$39$        & position held         & \texttt{<subject>  holds the position of <object>.}        & $2961$            &  $34835$       &   $\mathcal{D}^{\textsc{updated}}$    \\\hline
P$108$        & employer              & \texttt{<subject> works for <object>.}                  & $1544$               &  $20531$       &    $\mathcal{D}^{\textsc{updated}}$   \\\hline
P$102$        & political party       & \texttt{<subject> is a member of the <object>.}           & $1068$             &   $14232$      &   $\mathcal{D}^{\textsc{updated}}$    \\\hline
P$286$        & head coach            & \texttt{<object> is the head coach of <subject>.}        & $987$              &   $11935$      &   $\mathcal{D}^{\textsc{updated}}$    \\\hline
P$69$         & educated at           & \texttt{<subject> attended <object>.}        & $232$                          &    $2420$     &   $\mathcal{D}^{\textsc{updated}}$, $\mathcal{D}^{\textsc{unchanged}}$    \\\hline
P$488$        & chairperson           & \texttt{<object>  is the chair of <subject>.}          & $629$       &            $8468$     &   $\mathcal{D}^{\textsc{updated}}$    \\\hline
P$6$          & head of government    & \texttt{<object> is the head of the government of <subject>.} & $578$ &             $7815$     &   $\mathcal{D}^{\textsc{updated}}$    \\\hline
P$279$        & subclass of           & \texttt{<subject> is a subclass of <object>.}                 &            $5$ & $70$      &   $\mathcal{D}^{\textsc{new}}$, $\mathcal{D}^{\textsc{updated}}$    \\\hline
P$127$        & owned by              & \texttt{<subject> is owned by <object>}.                      &    $394$ &        $5326$      &   $\mathcal{D}^{\textsc{updated}}$, $\mathcal{D}^{\textsc{unchanged}}$    \\\hline
P$1001$       & legal term            & \texttt{<subject> is a legal term in <object>}.               &     $37$ &      $423$       &     $\mathcal{D}^{\textsc{unchanged}}$  \\\hline
P$106$        & profession            & \texttt{<subject> is a <object> by profession.}               & $83$ &           $1090$      &    $\mathcal{D}^{\textsc{updated}}$, $\mathcal{D}^{\textsc{new}}$, $\mathcal{D}^{\textsc{unchanged}}$   \\\hline
P$27$         & citizen               & \texttt{<subject> is <object> citizen.}                       &    $147$ &        $1983$      &   $\mathcal{D}^{\textsc{new}}$, $\mathcal{D}^{\textsc{unchanged}}$    \\\hline
P$176$        & produced by           & \texttt{<subject> is produced by <object>.}                   &   $24$ &         $276$      &   $\mathcal{D}^{\textsc{new}}$, $\mathcal{D}^{\textsc{unchanged}}$     \\\hline
P$138$        & named after           & \texttt{<subject> is named after <object>}.                   &    $73$ &        $1009$      &    $\mathcal{D}^{\textsc{new}}$, $\mathcal{D}^{\textsc{unchanged}}$    \\\hline
P$937$        & work location         & \texttt{<subject> used to work in <object>}.                  &     $38$ &        $507$     &    $\mathcal{D}^{\textsc{new}}$, $\mathcal{D}^{\textsc{unchanged}}$    \\\bottomrule
\end{tabularx}
}}
\caption{The list of templates we used for each relation in the \dtemplama{} dataset.}
\label{table:full_relations}
\end{table*}

\begin{table*}[t]
\small
\centering
\resizebox{\textwidth}{!}{
{\renewcommand{\arraystretch}{1.6}
\begin{tabularx}
{\linewidth}{
    >{\hsize=0.5\hsize}X
    >{\hsize=0.5\hsize}X
    >{\hsize=1.4\hsize}X
    >{\hsize=1.4\hsize}X
    >{\hsize=0.4\hsize}X
  }
\toprule
\textbf{\textsc{Wikidata ID}} & \textbf{\textsc{Relation}  }                                                                                       & \textbf{\textsc{Input} }& \textbf{\textsc{Labels}} & \textbf{\textsc{Split}} \\\hline\hline
P$54$         & member of sports team &    \texttt{Cristiano Ronaldo plays for \_X\_.}         &     \texttt{Juventus F.C., Manchester United F.C.}                       &  \texttt{2021-Q3}     \\\hline
P$39$        & position held         &  \texttt{Martina Anderson holds the position of \_X\_.}      &  \texttt{member of the European Parliament}  & \texttt{2019-Q4	}   \\\hline
P$108$        & employer              &   \texttt{George van Kooten works for \_X\_.}    &\texttt{University of Cambridge}     & \texttt{2022-Q2} \\\hline
P$102$        & political party       &   \texttt{Elena Kountoura is a member of the \_X\_.}   &   \texttt{Independent Greeks, SYRIZA} & \texttt{2019-Q2	}  \\\hline
P$286$        & head coach            &    \texttt{\_X\_ is the head coach of New York Red Bulls.}   &   \texttt{Gerhard Struber} & \texttt{2020-Q4}    \\\hline
P$69$         & educated at           &  \texttt{Sarafina Nance attended \_X\_.}    &   \texttt{Tufts University, University of California, Berkeley}   & \texttt{2020-Q2} \\\hline
P$488$        & chairperson           &  \texttt{\_X\_ is the chair of Lloyds Banking Group.}    & \texttt{Lord Blackwell} &  \texttt{2022-Q2}    \\\hline
P$6$          & head of government    &   \texttt{\_X\_ is the head of the government of United Kingdom.}   & \texttt{Theresa May, Boris Johnson} & \texttt{2019-Q3}     \\\hline
P$279$        & subclass of           &   \texttt{Mercedes-Benz A-Class is a subclass of \_X\_.}    &  \texttt{compact car} & \texttt{2022-Q2}    \\\hline
P$127$        & owned by              & \texttt{DeepMind is owned by \_X\_.}      &  \texttt{Alphabet Inc.} & \texttt{2021-Q4}     \\\hline
P$1001$       & legal term            & \texttt{Commonwealth of Independent States Free Trade Area is a legal term in \_X\_.}      &  \texttt{'Ukraine',
 'Russia',
 'Belarus',
 'Armenia',
 'Kazakhstan',
 'Moldova',
 'Kyrgyzstan',
 'Uzbekistan',
 'Tajikistan'}  & \texttt{\texttt{2022-Q2}}   \\\hline
P$106$        & profession            &   \texttt{Penny James is a \_X\_ by profession.}   & \texttt{chief executive officer}   & \texttt{2019-Q3}  \\\hline
P$27$         & citizen               &   \texttt{Yulia Putintseva is \_X\_ citizen.}   &  \texttt{Kazakhstan} \texttt{2022-Q1	}     \\\hline
P$176$        & produced by           &  \texttt{Land Rover Discovery series is produced by \_X\_.}     & \texttt{Jaguar Land Rover} & \texttt{2022-Q2	}      \\\hline
P$138$        & named after           &   \texttt{Bayes Business School is named after \_X\_.}   &   \texttt{Thomas Bayes} & \texttt{2021-Q3}    \\\hline
P$937$        & work location         &  \texttt{Eliza Vozemberg used to work in \_X\_.}    &    \texttt{Strasbourg, City of Brussels} & \texttt{2022-Q2}   \\\bottomrule
\end{tabularx}
}}\caption{Examples of \dtemplama{} for each relation.}
\label{table:full_relations_examples}
\end{table*}

\subsection{Full Results}
\begin{figure*}[t!]
    \begin{subfigure}{\textwidth}
        \centering
        \includegraphics[width=0.9\textwidth]{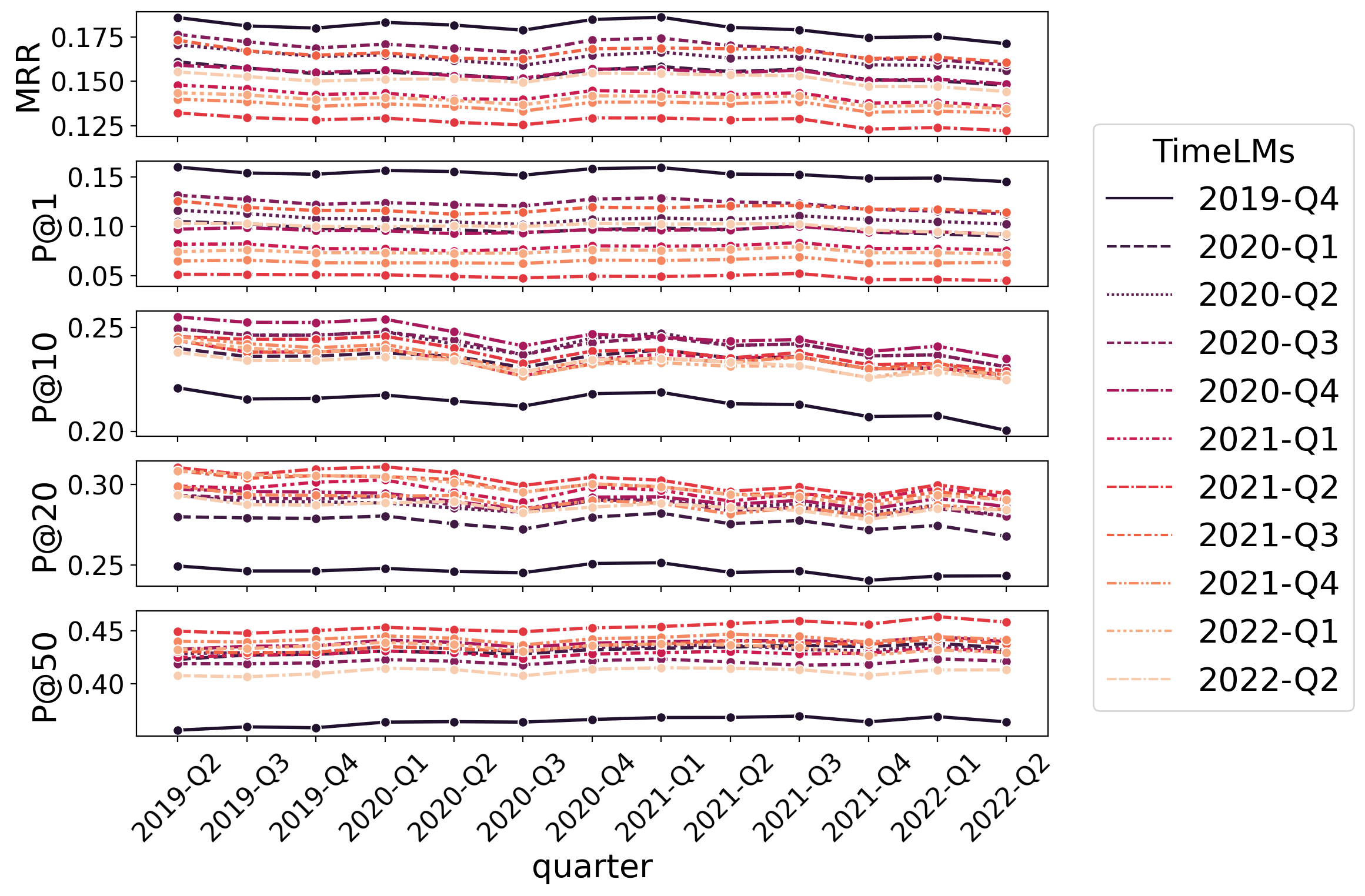}
        \caption{Single Token}
    \end{subfigure}%
    \\[\baselineskip]
    \begin{subfigure}{\textwidth}
        \centering
        \includegraphics[width=0.9\textwidth]{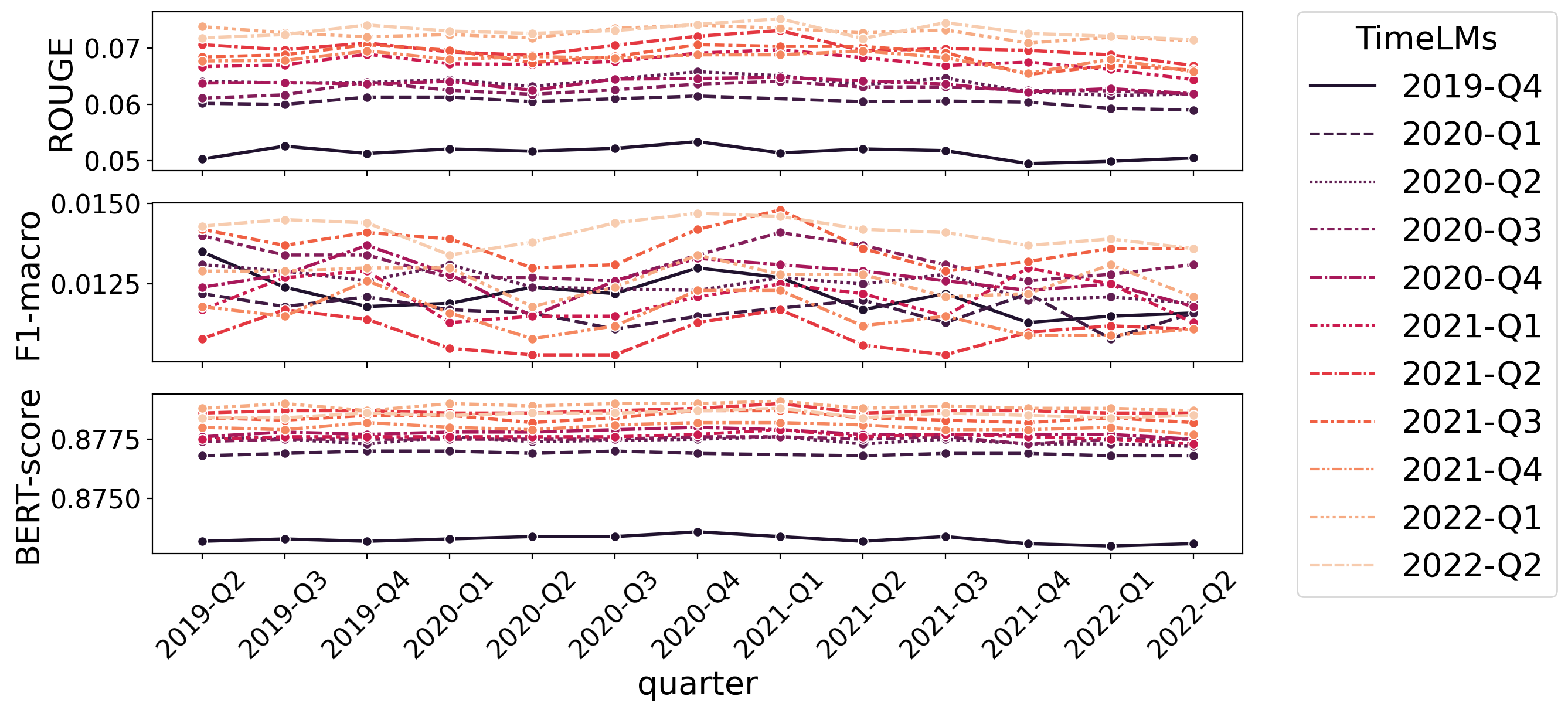}
        \caption{Multi-token}
    \end{subfigure}%

    \caption{Single-token probing and multi-token generation for the \textsc{Unchanged} split.
    }
    \label{fig:unchanged_full}
\end{figure*}

We provide the full results with all metrics for the \textsc{Unchanged} split in Figure~\ref{fig:unchanged_full}, and the \textsc{Updated}, \textsc{New} and \textsc{Deleted} splits for multi-token generation in Figure~\ref{fig:multi_token_full}.

\begin{figure*}[t!]
    \begin{subfigure}{\textwidth}
        \centering
        \includegraphics[width=0.9\textwidth]{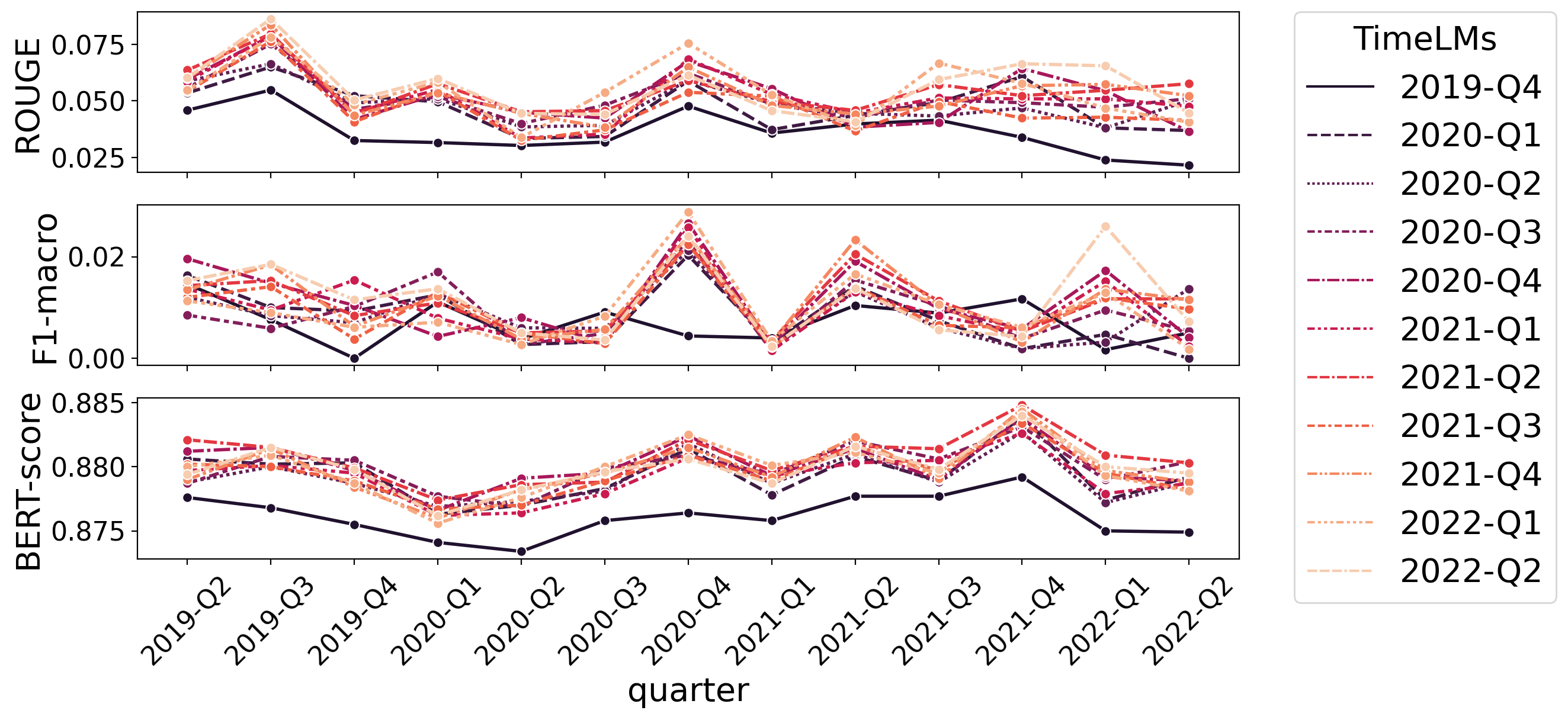}
        \caption{\textsc{Updated} Split}
    \end{subfigure}%
    \\[\baselineskip]
    \begin{subfigure}{\textwidth}
        \centering
        \includegraphics[width=0.9\textwidth]{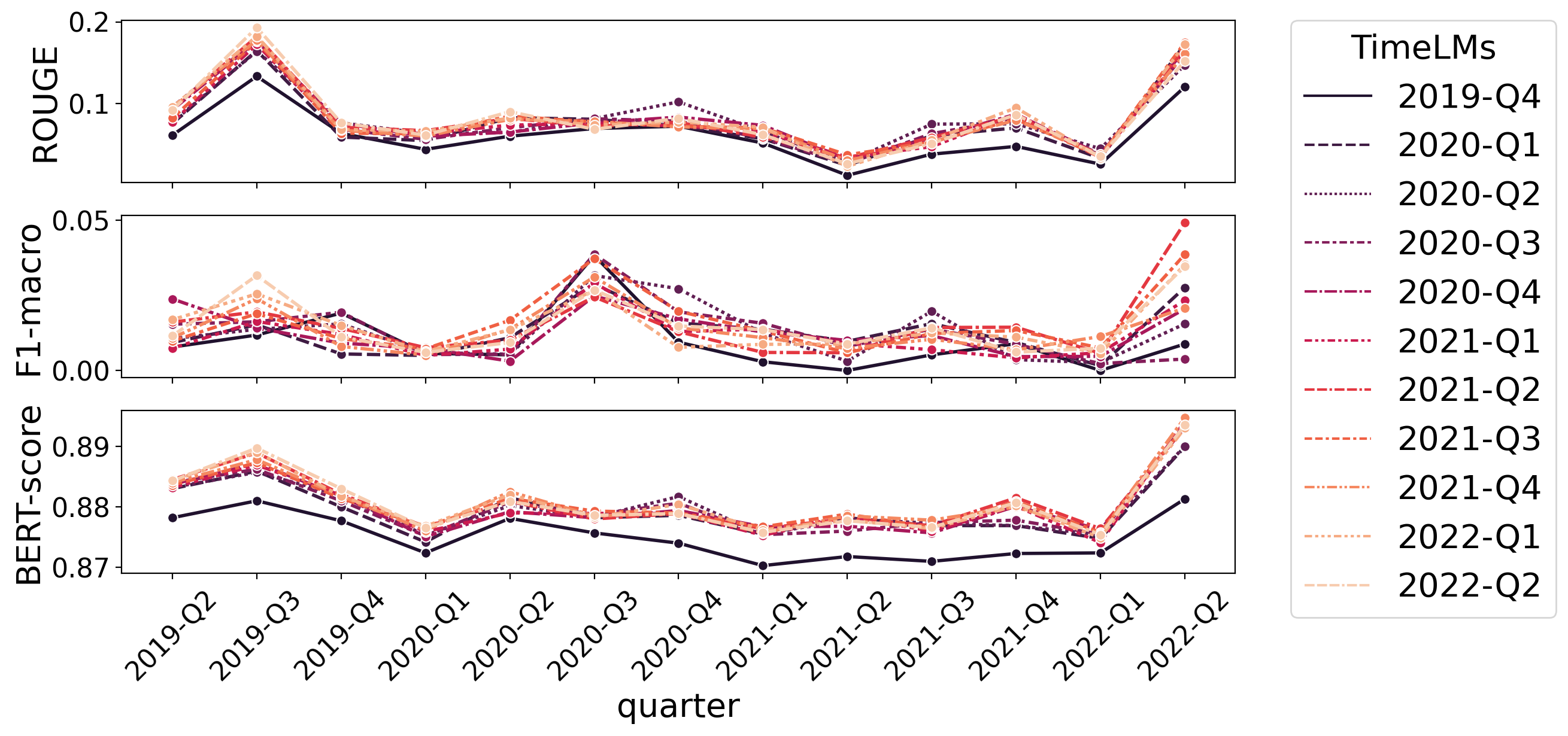}
        \caption{\textsc{New} Split}
    \end{subfigure}%
    \\[\baselineskip]
    \begin{subfigure}{\textwidth}
        \centering
        \includegraphics[width=0.9\textwidth]{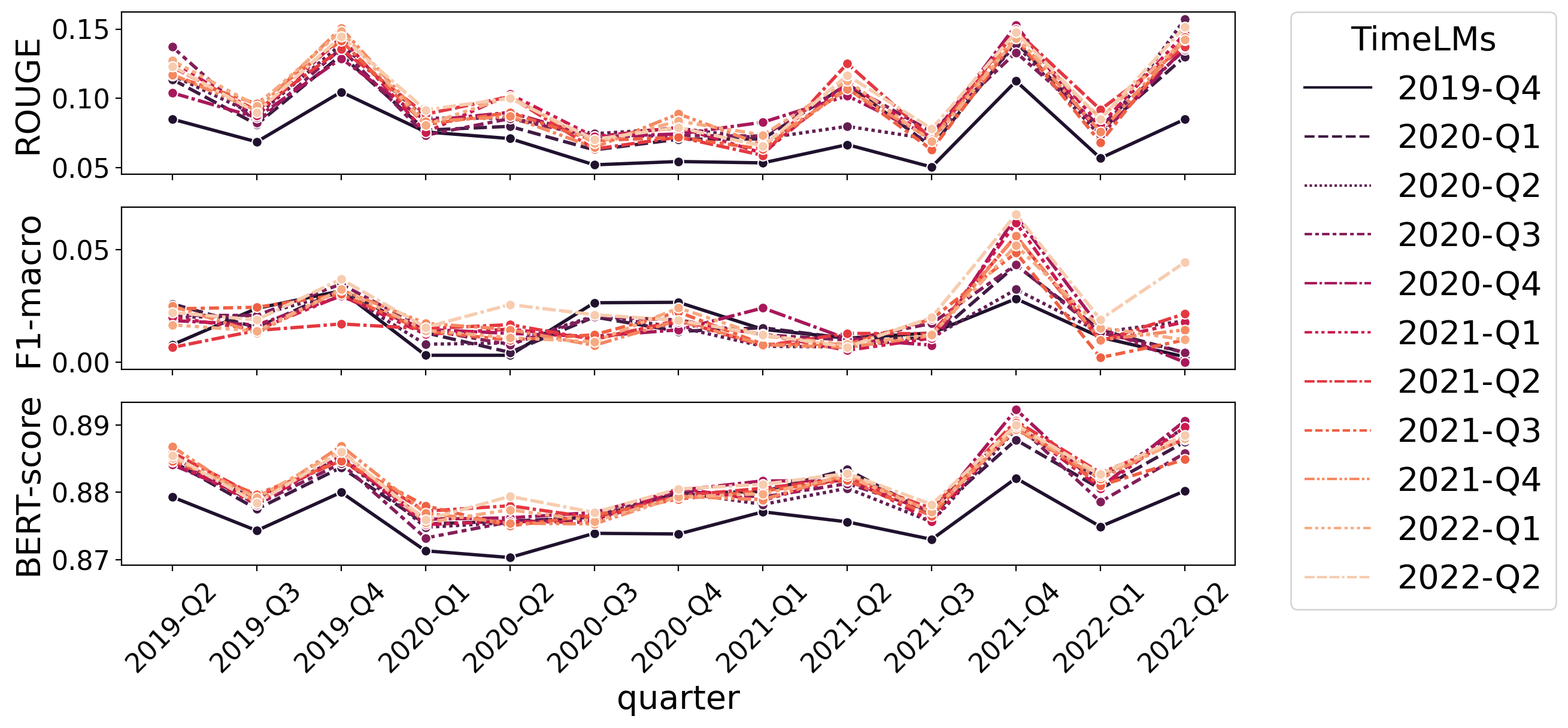}
        \caption{\textsc{Deleted} Split}
    \end{subfigure}
    \caption{Multi-token generation results for various fine-grained splits.
    }
    \label{fig:multi_token_full}
\end{figure*}

\begin{figure*}[t!]
    \begin{subfigure}{\textwidth}
        \centering
        \includegraphics[width=0.7\textwidth]{figures/updated_multi.png}
        \caption{\textsc{Updated} Split}
    \end{subfigure}%
    \\[\baselineskip]
    \begin{subfigure}{\textwidth}
        \centering
        \includegraphics[width=0.7\textwidth]{figures/new_multi.png}
        \caption{\textsc{New} Split}
    \end{subfigure}%
    \\[\baselineskip]
    \begin{subfigure}{\textwidth}
        \centering
        \includegraphics[width=0.7\textwidth]{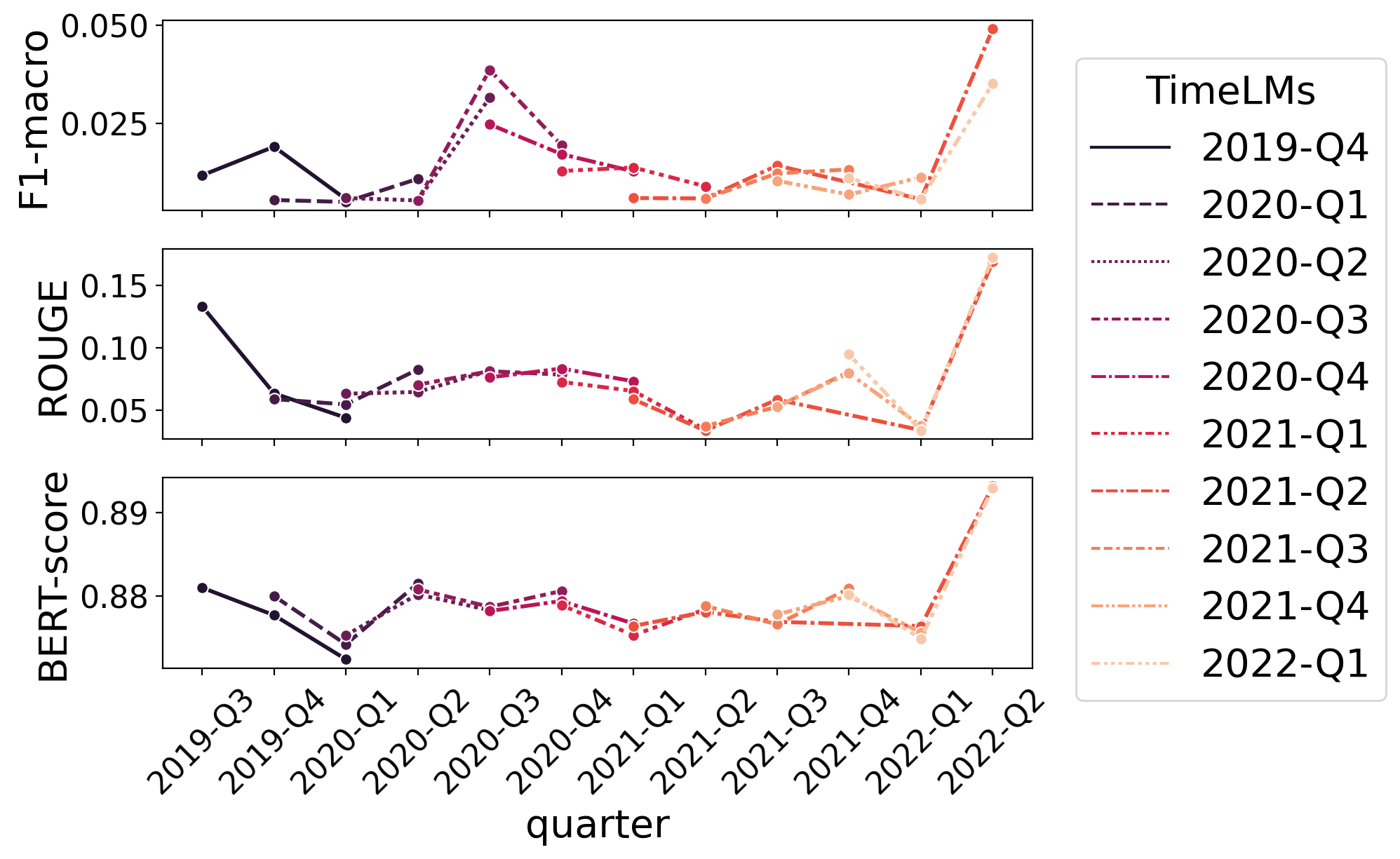}
        \caption{\textsc{Deleted} Split}
    \end{subfigure}
    \caption{Multi-token generation results for various fine-grained splits. Here for each model trained on timestep $t$, we keep the test sets from $t-1$, $t$ and $t+1$.
    }
    \label{fig:multi_token_full}
\end{figure*}
\end{document}